\documentclass{article}
\usepackage[utf8]{inputenc}
\usepackage{authblk}
\usepackage{setspace}
\usepackage[margin=1.25in]{geometry}
\usepackage{graphicx}
\usepackage{todonotes}
\usepackage{acronym}
\graphicspath{ {./figures/} }
\usepackage{subcaption}
\usepackage{threeparttable}
\usepackage{multirow}
\usepackage{multicol}
\usepackage{lineno}
\usepackage{graphicx} 
\graphicspath{ {images/} }

\usepackage[style=nejm, 
citestyle=numeric-comp,
sorting=none]{biblatex}
\title{An Interdisciplinary Outlook on Large Language Models for Scientific Research}

\author[1]{James Boyko}
\author[1]{Joseph Cohen}
\author[1]{Nathan Fox}
\author[2]{Maria Han Veiga}
\author[1]{Jennifer I-Hsiu Li}
\author[1]{Jing Liu}
\author[1]{Bernardo Modenesi} 
\author[1]{Andreas H. Rauch}
\author[1,*]{Kenneth N. Reid} 
\author[1]{Soumi Tribedi}
\author[1]{Anastasia Visheratina}
\author[1]{Xin Xie}

\affil[1]{University of Michigan - Michigan Institute for Data Science}
\affil[2]{Ohio State University - Department of Mathematics} 
\affil[ ]{All authors contributed equally to this work.}
\affil[*]{Address correspondence to: kenreid@umich.edu}

\date{}

\begin{document}
\newacro{AI}[AI]{artificial intelligence}
\newacro{DL}{deep learning}
\newacro{GAI}[GAI]{generative artificial intelligence}
\newacro{IP}[IP]{intellectual property}
\newacro{IR}[IR]{information retrieval}
\newacro{LLM}[LLM]{Large Language Model}
\newacro{ML}[ML]{machine learning}
\newacro{NAS}{neural architecture search}
\newacro{NLP}{natural language processing}
\newacro{SA}{sentiment analysis}
\newacro{XAI}{explainable artificial intelligence}
\newacro{HCI}{Human-Computer Interaction}
\newacro{EEB}{Ecology and Evolutionary Biology}
\newacro{IDE}{integrated development environment}
\newacro{NN}{neural network}
\newacro{CV}{computer vision}
\maketitle

\begin{abstract}
In this paper, we describe the capabilities and constraints of Large Language Models (LLMs) within disparate academic disciplines, aiming to delineate their strengths and limitations with precision. We examine how LLMs augment scientific inquiry, offering concrete examples such as accelerating literature review by summarizing vast numbers of publications, enhancing code development through automated syntax correction, and refining the scientific writing process. Simultaneously, we articulate the challenges LLMs face, including their reliance on extensive and sometimes biased datasets, and the potential ethical dilemmas stemming from their use. Our critical discussion extends to the varying impacts of LLMs across fields, from the natural sciences, where they help model complex biological sequences, to the social sciences, where they can parse large-scale qualitative data. We conclude by offering a nuanced perspective on how LLMs can be both a boon and a boundary to scientific progress.
    
\end{abstract}

\section{Introduction}\label{sec:intro}

\acp{LLM} embody a class of artificial intelligence systems renowned for their massive training datasets, intricate neural network structures, and the advanced ability to mimic, generate, and fine-tune natural language. Developers train \acp{LLM} on extensive text corpuses that cover a wide array of linguistic styles, domains, and subjects. Throughout their training, \acp{LLM} cultivate a statistical grasp of language, which empowers them to produce coherent, context-aware text and tackle numerous natural language processing tasks. The substantial neural networks within \acp{LLM}, sometimes containing tens or hundreds of billions of parameters, enable them to discern subtle semantic links and generalize across varied language tasks effectively. The scale and architecture of \acp{LLM} enhance their capacity to encompass vast linguistic knowledge and demonstrate a type of artificial creativity reflective of their training data. \acp{LLM}’s proficiency in engaging with and crafting human-like text paves the way for a multitude of applications, from machine translation to creative writing. Simultaneously, they raise important considerations about their interpretability, ethical use, and the risks of potential misuse.

\acp{LLM} stand at the forefront of current discourse for their transformative potential in scientific research. In recent years, several high-profile \acp{LLM} such as GPT-4 \cite{openai2023gpt4}, BERT \cite{devlin2019bert}, and LaMDA \cite{thoppilan2022lamda} have made their mark, all of which rely on the innovative transformer architecture \cite{vaswani2017attention}. These models, pre-trained on sprawling text datasets, are adept at emulating human-like conversational text, fielding questions, aiding translations, crafting summaries, and executing a range of \ac{NLP} tasks with notable accuracy \cite{kasneci2023chatgpt}.

\acp{LLM} exhibit a remarkable capacity to assimilate data at scales and speeds unattainable for human researchers, redefining the bounds of knowledge consumption and application. The skill of pre-trained \acp{LLM} to produce and critique text based on straightforward prompts opens avenues for automating laborious components of the scientific process. For instance, \acp{LLM} can bolster manuscript development and computational endeavors with their generative power, expedite information retrieval and review, or create code for initial data analysis. This optimization of research processes allows scientists to devote more time to nuanced analysis and insights, fostering a shift towards more strategic and interpretative scientific engagement.

The potential benefits of \acp{LLM} bring to the fore a host of environmental, ethical, and scientific concerns. For instance, the environmental footprint of \acp{LLM} is significant, owing to their resource-intensive training processes. Ethically, these models risk propagating biases and misinformation due to their reliance on data sourced from the internet. Additionally, the intricate and stochastic nature of their training complicates issues of reproducibility, trustworthiness, and intellectual property rights. Tackling these issues is paramount for the responsible and advantageous deployment of \acp{LLM} within the research community and broader society.

Although \acp{LLM} are a relatively new phenomenon, they have already generated considerable debate across various academic disciplines. Figure 1 illustrates the rapid increase in publications over recent years (sourced from Scopus database searches using the criteria "large PRE/1 language PRE/1 model"). Yet, academic consensus on their research impact remains elusive. Recent literature captures a dichotomy of opinions: on one hand, some assert that \acp{LLM} are poised to transform research methodologies \cite{Friedman_2023}; on the other, skeptics argue that \acp{LLM} will not significantly alter the research landscape \cite{nytimesOpinionNoam}.

\begin{figure}[h]
    \centering
    \includegraphics[width=1\textwidth]{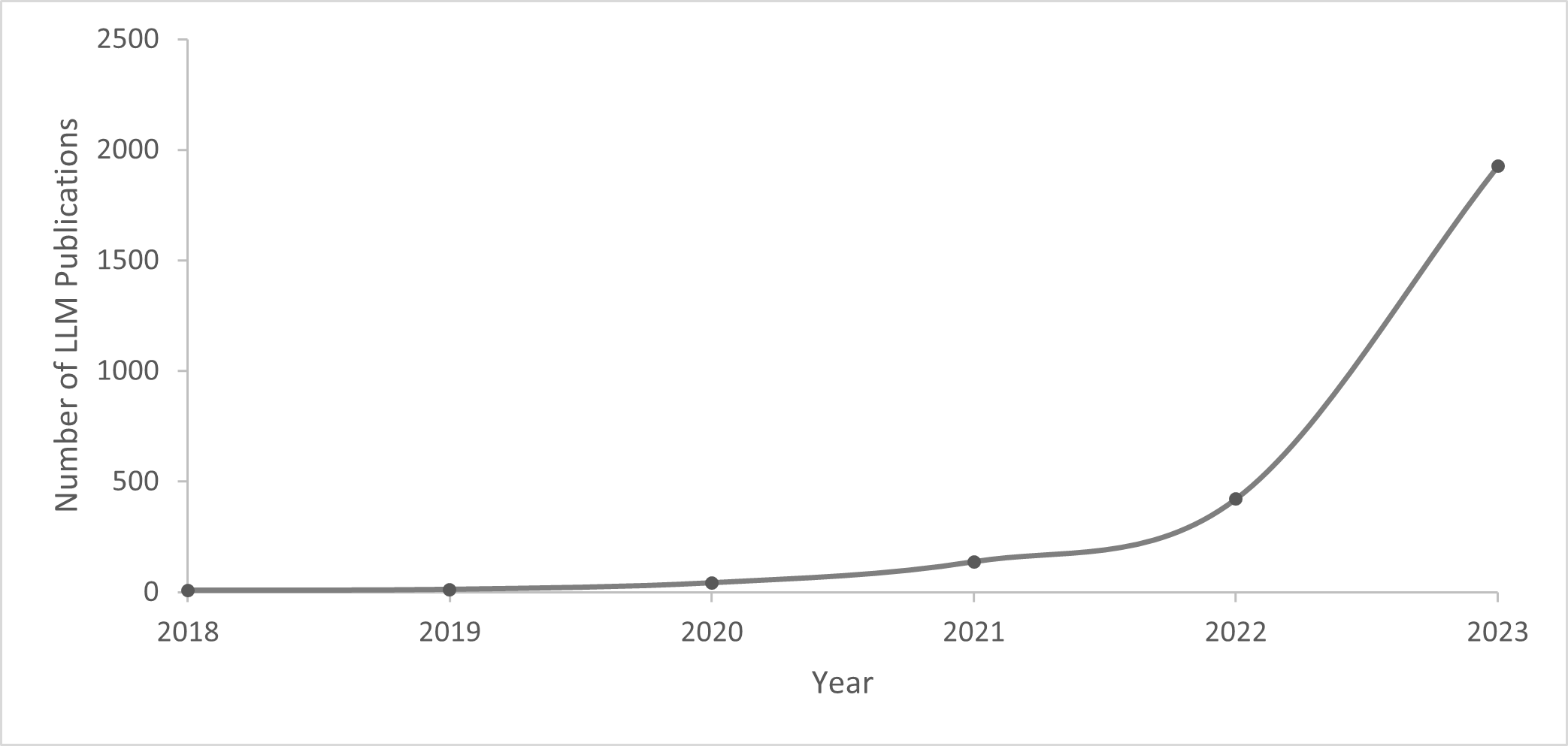}
    \caption{Number of Large Language Model publications. Scopus search for "large PRE/1 language PRE/1 model" within Article Title, Abstracts or Keywords. Search carried out on November 3\textsuperscript{rd} 2023.}
    \label{fig:mesh1}
\end{figure}

In this paper, we deliver a concerted evaluation of \acp{LLM} from the multidisciplinary perspectives of Schmidt AI Fellows, Data Science Fellows, and Research Staff at the Michigan Institute for Data Science. With our collective expertise spanning over 10 diverse fields\footnotemark{}, we endeavor to shed light on the utility of \acp{LLM} in broad research activities and their specific implementation in distinct scientific domains. Concurrently, we underscore the varying efficacy and potential limitations of \acp{LLM} across these disciplines. This document captures the contemporary state of \acp{LLM} at a time when the research environment is swiftly transforming, providing a timely overview of the current advantages, obstacles, and our collective viewpoint.

\footnotetext{The science domain backgrounds of the authors are, JB: Ecology and Evolutionary Biology, JC: Mechanical Engineering, NF: Environmental Science, MHV: Mathematics, JIL: Astronomy, JL: Genetics and Developmental Biology, BM: Economics and Statistics, AHR: Aerospace Engineering, KNR: Computer Science, ST: Chemistry, AV: Nanoscience \& Nanotechnology, and XX: Physics.}

In terms of terminology to describe \acp{LLM}, general \textit{pre-trained} models, like GPT-4, have garnered significant mainstream attention, however, these models are just one class of \ac{LLM}. Another class of models consists of pre-trained \acp{LLM} that are specialized on a domain by further training on domain specific data. Such models will be referred to as \textit{fine-tuned} \acp{LLM} throughout the paper. Further, the transformer architecture, as the basis of \acp{LLM}, has been used to great success with sequential, but not necessarily textual, data in fields as diverse as biology, chemistry, economics and health sciences. We refer to these models as \textit{domain-specific} \acp{LLM}. This work will discuss all three of these \acp{LLM} classes and their potential impact on scientific research. First we will discuss general research tasks before exploring domain-specific applications and the future outlook for \acp{LLM} in research.

\section{Research Tasks: Applications and Challenges}\label{sec:researchtasks}

\acp{LLM} provides great opportunities in promoting the research workflows across scientific domains. Meanwhile, there still exist limitations and challenges that need to be noticed when \acp{LLM} are applied to research methodologies. Here we summarize applications and challenges of \acp{LLM} in general research workflows across scientific domains. As these research tasks are common across all fields of research, the discussion in this section focuses on general pre-trained and fine-tuned \acp{LLM}. The next section (Sec. \ref{sec:domainspecifictasks}) explores domain-specific models.

\subsection{Applications} \label{sec:researchtasks:subsec:Applications}
The potential for \ac{LLM} integration in research workflows across scientific domains is immense. These opportunities will be summarized similarly to the categories identified by \cite{korinek2023} and are ordered to reflect a general scientific research workflow: ideation and information review, coding and data analysis, and writing. For each research task, we review how \acp{LLM} can be used for each stage of the scientific process, specifically: ideation and information review in section \ref{sec:researchtasks:subsec:ideationandinformation}, coding in section \ref{sec:researchtasks:subsec:coding} and writing in section \ref{sec:researchtasks:subsec:writing}, as well as some of the challenges associated with each task. Although challenges are mentioned briefly in each subsection, major challenges are described in detail in section \ref{sec:researchtasks:subsec:concerns}. 

\subsubsection{Ideation and Information Review} \label{sec:researchtasks:subsec:ideationandinformation}

Traditionally, information is gathered manually from the literature. Information review is the process of obtaining relevant information from a large collection of unstructured or semi-structured data, and is essential to research design and ideation. In many research domains, information is spread across unstructured text, figures, and tables in a huge body of research papers. While existing \ac{IR} systems such as Google Scholar, PubMed, and arXiv streamline the process of conducting thorough literature reviews, information overload remains an outstanding problem in most academic domains \cite{Nyamisa2017ASO}. Nonetheless, \acp{LLM} have already shown to have excellent performance on information retrieval and review tasks due to being pre-trained on a large corpus of data. \acp{LLM} can parse through texts, papers, and other natural language sources while extracting relevant information and relations at a pace no human can match. Potentially, these capabilities can lead to significant accelerations in the generation of research questions, summarization of existing literature, and identification of research gaps. 

In addition, the potential for \acp{LLM} to automatically extract data from text sources could lead to an influx of meta-analytic data at an unprecedented scale. This can include both textual and tabular data \cite{yue2023leveraging, Jeronymo2023InParsv2LL} and has been successfully applied in many diverse domains such as economics, aviation safety, and health settings \cite[e.g.,][]{zhu2023chatgpt, SafeAeroBERT, aeroBERT, FineTuningLLM, astroBERT}. We will further showcase some fine-tuned and domain-specific \acp{LLM} in Section \ref{sec:domainspecifictasks}. Generally, these frameworks aim to automate the synthesis and summarization of findings across multiple studies, enabling researchers to gain valuable insights and identify patterns in the collected data.

Recent research states found that \acp{LLM} can now generate research questions \cite{agathokleous2023use}. While current use has centered on general models, domain-specific \acp{LLM} are emerging, which will better integrate and apply to growing domain literature. These models promise to distill domain knowledge, sparking new research directions. Additionally, \acp{LLM} can foster interdisciplinary work by bridging knowledge gaps across scientific fields.

\subsubsection{Coding} \label{sec:researchtasks:subsec:coding}

\acp{LLM} excel in many coding tasks such as code generation, completion, synthesis and translation \cite{LLMCode-Xu}. From simple text prompts, \acp{LLM} can generate and rewrite code with high readability and efficiency. The ubiquity of coding tasks in research means that these capabilities may be a massive aid for academics who need to conduct simulations, data processing, graphical visualization, or data analysis. The near instantaneous generation of code, combined with recursive feedback from users where any errors messages encountered are fed back into the \ac{LLM}, can lead to massive time savings compared to a manual search of documentation or question-and-answer websites. Common coding errors, such as syntax errors, incorrect function usage or object instantiation, can now rapidly be solved with a few text prompts to a general pre-trained LLM. In the past this would require tedious line-by-line code checking or, as compared to prompting an LLM, more time spent searching for answers in code forums and online code documentation. Vaithilingam et al. \cite{Vaithilingam-chi-2022} evaluated the usability of \acp{LLM} for code generation, and found that it is an excellent tool to kick-start a task without requiring the effort of searching online. These sorts of time and effort savings are especially significant when using new programming languages, or for researchers who do not have extensive programming experience. The lack of domain specificity of many coding tasks will heighten the usefulness of \acp{LLM} across a broad range of disciplines. To that end, examples of ChatGPT generating code already exist for a diverse set of tasks. Code for modeling carbon concentrations \cite{zhu2023chatgpt}, molecular analysis \cite{Glen2022, white_future_2023, pimentel}, and even solving partial differential equations \cite{Kashefi2023} have all been generated via ChatGPT without additional fine-tuning. Furthermore, \acp{LLM} are just as impressive at translating between coding languages as they are for generating new code from text prompts \cite{roziere-neurips-2020,kulal-neurips-2019}, enabling researchers to quickly and efficiently modernize existing code within their field. 

\ac{LLM}-based coding assistants can thus greatly streamline the code development pipeline. A useful \ac{LLM}-based coding assistant is GitHub Copilot, which has integrated Codex, a \ac{LLM} trained specifically for coding tasks, into an \ac{IDE}. Nonetheless, caution is needed as these models can generate bugs or security vulnerabilities that may be difficult to discern, especially when generating many lines of code \cite{liu-arxiv-2023}. Tools such as Codex should be used with caution as powerful yet imperfect coding assistants. Another general concern of \acp{LLM} is reproducibility. Due to the stochastic nature of \acp{LLM} the same text prompt may generate different code, users should bear this in mind and check \acp{LLM} generated code for correctness and completeness. Nonetheless, even with errors, it is generally faster to code with \ac{LLM} assistance, especially for unfamiliar coding tasks. 

\subsubsection{Writing} \label{sec:researchtasks:subsec:writing}

Effective communication of scientific results is essential for advancing research, securing funding, and informing the public; it hinges fundamentally on clear, precise writing. \acp{LLM} can be used to produce clear English statements based on a list of arguments and references, resembling advanced versions of grammar checking and editorial software \cite{Vishniac2023Editorial}. In addition, \acp{LLM} can prompt relevant ideas and topics during editing and rewriting, assisting authors to overcome writing blocks by presenting comprehensive perspectives. Using \acp{LLM} as a writing assistant is particularly beneficial for researchers unacquainted with scientific writing and non-native English speakers, as they can enhance writing style, composition, and punctuation, thus helping leveling the playing field. For public education and science communication purposes, \acp{LLM} can efficiently generate texts for diverse audiences, such as automated data summaries \cite{2023arXiv230500108C}, press releases highlighting new discoveries, and outreach articles tailored to various educational backgrounds \cite{agathokleous2023use}. Likewise, \acp{LLM} can facilitate the translation of scientific articles into multiple languages, thereby expanding the audience base.

\subsection{General Challenges}\label{sec:researchtasks:subsec:concerns}

Though \acp{LLM} have great promise for each stage of the scientific method, there are a number of limitations and challenges that need to be acknowledged by those looking to implement them in their own research. Challenges and limitations can be general, such as, explainability, reproducibility and environmental impact or primarily related to specific research tasks. Here, we highlight some of the key challenges and limitations users should be aware of when using \acp{LLM} for research.

With the increased popularity of \acp{LLM}, many scientific publishers have published updated policies on using generative AI in preparing scientific manuscripts. While some strictly prohibits any use of AI generated text \cite{science_editorial}, most journals allow \acp{LLM} as an editing tool for improving readability and languages \cite{AAS_editorial, ACS_editorial, ICML_editorial, nature_editorial, elsevier_editorial}, and producing new ideas and text solely based on \acp{LLM}, without critical review and editing by the human author, are generally considered scientific misconduct. As a tool/software, \acp{LLM} does not qualify for authorship per guidelines of all the above-mentioned journals, instead the use of \acp{LLM} should be disclosed and detailed in the methods or acknowledgments sections, similar to other software tools. However, with the fast evolution of \acp{LLM}, the policies regarding using \acp{LLM} in scientific research are likely to evolve with the technology. 

\subsubsection{Explainability}\label{sec:researchtasks:subsec:generalconcerns:subsubsec:explainability} 
Explainability, within the scope of neural networks and \ac{ML}, refers to the extent to which a model's internal processes and decision-making pathways can be elucidated and understood by humans. This attribute, generally known as \ac{XAI}, is an emerging sub-field within \ac{AI} research. An \ac{ML} model possessing high explainability allows not only for decomposing the effect of features on the outcome, but also for tracing how the model reached a particular result. Such transparent operation fosters trustworthiness, supports troubleshooting, enhances replicability, and assists in the tasks of bias detection and mitigation. In contrast, models with low explainability, often associated with complex \acp{LLM} like GPT-4 \cite{openai2023gpt4}, obfuscate the inner workings behind their outputs, resulting in a `black box' situation. 

Even though seminal methods for \acp{LLM} explainability have been recently developed, e.g. \cite{bills2023language, singh2023explaining, braşoveanu2022visualizing, li2022explanations}, challenges remain. The opaqueness of \acp{LLM}, which are effectively black boxes to users, largely stems from their complex architecture, in conjunction with the uncertainty of the inner mechanisms for generating output\footnote{The uncertainty associated from \acp{LLM} outputs stems from unclear source of training data, or unavailable training scripts and environment, in addition to the stochastic process of generating outputs. All of these factors impact both the ability to explain and reproduce results from these models.}. Transformer-based \acp{LLM} embed words into vectors, and process these through multiple nonlinear layers with `self-attention' mechanisms, then weighing the relevance of different parts of the input, and sampling from a probability distribution to produce text. Self-attention mechanisms allow a neural model to weigh and focus on different parts of the input based on their relevance to a given token, enabling the model to capture context-dependent relationships in sequences. Unfortunately, this context-based weighting system combined with stochastic sampling produces a somewhat opaque and non-inspectable algorithm, even with the use of state-of-the-art XAI tools. If a researcher is to inspect such a model with the intent of understanding how an output is generated, it would be required deep understanding of the mechanisms involved, in addition to a herculean task of combining the understanding of each submodel that compose the entirety \acp{LLM}. 

\acp{LLM} generate responses based on patterns in their training data; thus, biases or inconsistencies in their outputs often reflect the training data's nature. Lack of transparency in the training data complicates understanding the reasons behind a \ac{LLM}'s specific outputs and assessing its reliability across different domains. The undisclosed nature of the training data for many \acp{LLM} hinders explainability, posing a significant barrier to their ethical use in sensitive areas and in building public trust.

\subsubsection{Reproducibility}\label{sec:researchtasks:subsec:generalconcerns:subsubsec:reproducibility}
Reproducibility poses significant challenges to using \acp{LLM} as a central tool in research. The probabilistic sampling techniques and intricate training processes employed make obtaining consistent and reproducible results difficult. For example, given the same input and model state, the output will vary across runs due to the probabilistic sampling techniques used. However, reproducibility concerns stemming from this sampling procedure can be mitigated by adjusting model settings which regulate the randomness. For instance, a zero ``temperature'' setting makes GPT-3 output deterministic \cite{korinek2023}. Other aspects of reproducibility are more difficult to deal with. Model settings may be adjusted to control the output of identical prompts, but with even slight differences to the input prompt, such as punctuation, number of spaces, etc., divergent results may occur. Furthermore, \acp{LLM} are trained on distributed computing frameworks across multiple GPUs. The parallelization and synchronization parameters involved in distributed training are vast \cite{zhao2019dynamic}. Subtle differences in the learning process can lead to significant variations in the model's behavior and output. The computational scale required to train \acp{LLM} may make it infeasible to reproduce the exact training conditions or retrain the model from scratch. Finally, as with most software, \acp{LLM} are continuously updated and reproducing results from older model versions or comparing results across versions can be challenging if old versions are not maintained and available.

To address these challenges and promote reproducibility, making the training process of the \ac{LLM} model more transparent and more open source would be helpful. Meanwhile, researchers and practitioners are increasingly documenting the exact training and evaluation procedures, sharing code and models, providing explicit instructions for environment setup, and establishing standardized evaluation benchmarks. By establishing clear guidelines, open-sourcing codes, and promoting transparency in the both model developer and research community, \ac{LLM} reproducibility can be enhanced. 

\subsubsection{Privacy Concerns}\label{sec:researchtasks:subsec:generalconcerns:subsubsec:privacyconcerns}

While the ownership of data generated by \acp{LLM} is still unclear and yet to be settled in the courts, there's no ambiguity about the ownership of the data inputted into these models. This clarity contrasts sharply with another issue: a notable difference exists between \acp{LLM} functioning on standalone architectures and those integrated with wider networks. In the latter scenario, input data might be stored, examined, and potentially utilized for subsequent training by third parties. Popular \acp{LLM} privacy policies explicitly state that various data are captured during usage, for example, ChatGPT captures: \textit{log data, usage data, device information, cookies} and \textit{analytics} \cite{OpenAI_privacy}. Of particular import and concern for researchers is the statement ``We may automatically collect information about ... the actions you take''. Such a provision implies retention of all interactions with ChatGPT, encompassing facets like ideation, linguistic assistance, data interpretation, coding assistance, and more. Google's Bard `Privacy Help Hub' states ``To help with quality and improve our products (such as generative machine-learning models that power Bard), human reviewers read, annotate, and process your Bard conversations.'' \cite{GoogleBard}. While this is stated clearly in the privacy notice for ChatGPT and Bard, a substantial fraction of researchers will not delve into these privacy stipulations, whether due to a lack of awareness of the risks or a belief that the benefits of using \acp{LLM} outweigh the risks. Other \acp{LLM} may not be so clear with regard to stating the ownership of usage data. The ramifications are magnified when you consider data that is sensitive in nature. For instance, healthcare researchers working with confidential patient data, or researchers collaborating with government or industrial entities that involve propriety data of significant strategic or commercial value. 

However, in response to the burgeoning concerns over data privacy and the potential misuse of sensitive information, several mitigation strategies are being devised and implemented across the academic and industrial spectrum. The University of Michigan, recognizing the utility of \acp{LLM} yet cognizant of their inherent privacy vulnerabilities, has pioneered a novel approach with the introduction of `U-M GPT' \cite{um_gpt}. This model is exclusively accessible to UM personnel, thereby ensuring a controlled environment and substantially reducing the risk of unintended data exposure of research related data and prompts to external entities. `U-M GPT' collects personal information for operational purposes, doesn't use user data to train its AI models, doesn't sell or license personal information, shares data with specific service providers (i.e. Microsoft Azure) for operational reasons, and implements measures to protect data, while being subject to periodic updates in its privacy policy (accurate as of publication of this article). Similarly, recognizing the need for enhanced data protection, OpenAI has announced an enterprise version of ChatGPT \cite{chat_gpt_enterprise}. This adaptation offers privacy guarantees, ensuring that sensitive information remains within the confines of the designated enterprise. There are also an increasing number of institutions and businesses that are opting for local installations of \acp{LLM}. By hosting models on their own infrastructure, they effectively sidestep the risks associated with cloud-based or third-party managed solutions. Such local deployments grant organizations full autonomy over data management, allowing them to institute data protection protocols in line with their specific requirements and the nature of the data they handle. Through these measures there are attempts to harness the potential of \acp{LLM} while concurrently addressing the pivotal issue of data security. This consideration can be found in current medical literature for dealing with sensitive data \cite{bumgardner2023local}, social sciences for community based data pooling while retaining individual and community data security \cite{southsecure}.

\subsubsection{Environmental Impact}\label{sec:researchtasks:subsec:generalconcerns:subsubsec:environmentalimpact}
\acp{LLM} have both direct and indirect impacts on the environment. Training \acp{LLM} requires extensive computational resources, including powerful processors and high-capacity storage systems \cite{carbontracker_2020}. These computations consume a substantial amount of electricity, which often comes from fossil fuel-based power generation, contributing to carbon emissions. Luccioni estimates the power consumption of training various \acp{LLM}, all with billions of parameters, to be in the range of $324$ to $1287$ MWh resulting in carbon emissions in the range of $30$ to $552$ tons of CO$_2$ \cite{luccioni_estimating_2022}. The authors further highlight the impact that local data center power usage efficiency and energy source has on carbon emissions. For example, a \ac{LLM} trained in the central US would emit an order of magnitude more CO$_2$ than a \ac{LLM} trained in France. Additionally, popular \acp{LLM} such as ChatGPT have millions of active users, resulting in an operational electricity cost that far exceeds training costs \cite{reuters_gpt_users}. Maintaining and upgrading \acp{LLM} require constant hardware updates. The rapid development of more powerful hardware can render existing equipment obsolete, leading to electronic waste disposal concerns. To promote sustainability, it is important to consider the environmental implications of using \acp{LLM}. Before selecting an \acp{LLM} for use in your specific field, consider:

\begin{enumerate}
    \item Utilizing smaller models when possible, as they often require less energy.
    \item Leveraging existing pretrained models rather than training new ones from scratch.
    \item Being conscious of the location and energy sources when using computational resources.
\end{enumerate}

By choosing more environmentally-friendly approaches, scientists can help drive demand in a direction that is less taxing on our planet. Being informed and making sustainable choices can ensure that the benefits of \acp{LLM} are enjoyed without undue environmental costs.

\subsubsection{Misinformation}\label{sec:researchtasks:subsec:generalconcerns:subsubsec:misinformation}
\acp{LLM} are trained on vast amounts of data, often from unknown sources via the internet, which can lead to misinformation by absorbing the biases and inaccuracies present in their training data \cite{politicalbias,biasneurips}. This absorption is not limited to factual inaccuracies, but also extends to ethical and moral norms, reflecting the societal biases present in the data. This issue is further enforced by the fact that \acp{LLM} currently lack the ability to discern the veracity of the information processed. Generally, they do not have a built-in mechanism to fact-check or validate the information against reliable sources. As a result, an \ac{LLM} might inadvertently spread misinformation, especially if it is prevalent in the training data. This is particularly concerning given the increasing reliance on \acp{LLM} for tasks such as document retrieval, sentence selection, and claim verification, where the accuracy and reliability of the information are paramount \cite{nie2019combining}. Misinformation from \acp{LLM} could potentially misguide researchers, negatively impacting the progress of research by creating false paths for other scientists to follow. 

\subsubsection{Hallucinations}\label{sec:researchtasks:subsec:generalconcerns:subsubsec:hallucinations}
In addition to generating misinformation, as discussed in section \ref{sec:researchtasks:subsec:generalconcerns:subsubsec:misinformation}, \acp{LLM} are also capable of generating information that is entirely fictional when asked about information outside its training data, resulting in what is often described as hallucinations. The presence of hallucinations  poses serious challenges for scientific research \cite{banghallucination,alkaissihallucination}. Researchers who rely on language models to assist in literature reviews will find that \acp{LLM} refers to non-existent citations, leading to potential confusion, wasted effort, and the danger of dissemination of misinformation. When it comes to using \acp{LLM} to explain theories and concepts, we have also faced incorrect explanations and interpretations of scenarios. Although a powerful tool, these models are not currently able to provide completely factual information and therefore researchers will need to critically evaluate and validate \ac{LLM}-generated information, cross-referencing it with established knowledge and authoritative sources. Existing models which are not integrated with the internet are unable to provide accurate citations for the content that they generate. However, this limitation is being overcome by state-of-the-art models such as Bard \cite{bard2023} and Elicit \cite{elicit} which are able review and summarize literature directly while providing direct sources.

\subsubsection{Plagiarism and Intellectual Property}\label{sec:researchtasks:subsec:generalconcerns:subsubsec:plagiarismandintellectualproperty}

Plagiarism and \ac{IP} issues of \acp{LLM} revolve around the concern of unauthorized content reproduction or generation that may infringe upon established copyright laws, trademark protections, and other forms of \ac{IP} rights. Being trained on large-scale datasets, \acp{LLM} often encompass numerous works that are subject to \ac{IP} rights. As \acp{LLM} generate outputs based on their learned patterns and structures from these datasets, there exists a non-negligible probability that their generated content could unintentionally mimic or replicate protected works, thus inadvertently plagiarizing. \ac{LLM}'s ability to create novel content engenders ambiguity concerning who owns the \ac{IP} rights to such generated outputs. The delineation between what constitutes a transformative use, permissible under fair use doctrine, versus explicit infringement remains a gray area in \ac{AI} law. Consequently, these looming \ac{IP} challenges pose profound implications for the governance and regulatory landscape of \acp{LLM}, necessitating urgent scholarly attention and legislative action. Currently, OpenAI is facing a lawsuit over how it used people's data to train ChatGPT \cite{cnn-news-article}. Ultimately, \acp{LLM} should be treated as a tool in aiding the writing process in the research pipeline, and the human authors should be held accountable for any statements or ethical breaches.

\section{Domain-Specific Research Tasks}\label{sec:domainspecifictasks}
Effective application of language models for addressing domain-specific research comes down to one question: can data in that domain be meaningfully represented as a sequence? Language entails a structured sequence of words such that an overall meaning can be attached to it. In some domains, data are naturally sequentially structured and as a consequence, in such areas of research there have been a plethora of \ac{LLM} applications in the past two to three years. In this section, we will discuss some recent applications of \ac{LLM} architecture in solving research problems, sectioned into various domains. This section discusses domain-specific \acp{LLM}, models trained on non-textual data in a specific domain, as well as general or fine-tuned \acp{LLM} where appropriate.

\subsection{Biological Sciences}

For centuries, biological knowledge has been disseminated through the scientific literature. Vast archives of papers in scientific journals and preserved specimens in natural history museums contain invaluable information on biological diversity. However, this information has proven to be inaccessible for large data analysis \cite{Lffler2020DatasetSI} and its automatic extraction has been historically limited \cite{Thessen2012ApplicationsON}, thus hampering research into ongoing human-caused biodiversity crises. The efficient access and extraction of this information would give scientists access to a vast amount of data. With the remarkable improvements of \acp{LLM} to comprehend text, there is potential for the automated annotation of taxonomic texts with the eventual extraction of morphological character information\cite{Lcking2021MultipleAF, Nguyen2019COPIOUSAG,Thessen2018AutomatedTE, Mora2018SemiautomaticEO}. However, these phenotypic descriptions are often difficult to mine with pre-\ac{LLM} technologies because of homonymy and synonymy of morphological descriptions \cite{Balhoff2013ASM}. Although, the automatic extraction of morphological, taxonomic, and geospatial information from biological text \cite{Lcking2021MultipleAF, Nguyen2019COPIOUSAG,Thessen2018AutomatedTE, Mora2018SemiautomaticEO} has seen active development in recent years and the integration of foundational \acp{LLM} into these toolkits has the potential to further revolutionize and accelerate biological sciences towards an even more data-driven science.

\subsection{Chemical Sciences}

Multiple language models have been shown to perform quite well in property prediction with string-based representations. For example, \acp{LLM} like SMILES-BERT \cite{wang_smiles-bert_2019}, ChemBERTa-2 \cite{ahmad_chemberta-2_2022}, which was based on RoBERTa, and MoLFORMER \cite{ross_large-scale_2022} developed by IBM show that language models perform quite well at learning molecular properties solely from string-based short representations as compared to other \ac{DL} methods. Language models have also been shown to perform better than some graph generative models in generating complex molecular structures with specified characteristics \cite{flam-shepherd_language_2022}.

\subsection{Engineering}

As \acp{LLM} improve capabilities for applications in fundamental sciences, they will likely be useful for engineering applications in the future as well. At the time of this writing, \ac{LLM} integration into engineering research remains sparse and somewhat limited in scope to the general tasks outlined in the previous section. As a result, \acp{LLM} have not yet made a significant impact on engineering problem-solving. Engineering problems often require careful application of mathematics, statics, dynamics, chemistry, principles from other domains, and specific domain experience that may not be well-documented or represented in training corpora. 

Despite these limitations, there is significant untapped potential for \acp{LLM} to positively impact engineering disciplines. Preliminary studies on \ac{LLM} performance on the Fundamentals of Engineering (FE) exam, a standardized licensing test for professional engineers, have shown that models such as GPT-4 can achieve an overall accuracy of around 75\% on the Environmental exam with minimal prompt modifications \cite{pursnani2023performance}. Considering the passing score of 70\% and the overall pass rate of 64\% \cite{pursnani2023performance} this indicates the potential \acp{LLM} have in tackling engineering problems.

 The existing research in engineering has positioned \acp{LLM} as a tool to optimize workflows and improve productivity. For example, compliance with federal regulations is essential for safety-critical applications, but reviewing these guidelines can be time-consuming and labor-intensive. Fine-tuned BERT models such as aeroBERT \cite{aeroBERT} and SafeAeroBERT \cite{SafeAeroBERT} have already been researched for their potential usage in reviewing safety regulations for the aerospace industry. \ac{LLM}-enabled code generation may prove instrumental for yielding process optimization benefits in additive manufacturing, where Gcode is used for controlling layer-by-layer printing and process parameters \cite{badini2023assessing}. A further potential application of \acp{LLM} in engineering is in aiding causal reasoning tasks.  \cite{kıcıman2023causal} uses \acp{LLM} to assess the arrow of causality between cause-effect pairs such as the rotation of a Stirling engine and the heat bath temperature as well as the cement ratio and the compressive strength of concrete. While these cause-effect pairs are applied engineering, rather than research, this shows how \acp{LLM} can translate engineering domain knowledge into formal methods for downstream causal reasoning tasks, and is thus an exciting opportunity for enhancing research on causal effects in engineering.
 
 Ultimately, the potential for \acp{LLM} in addressing fundamental problems in engineering will depend on how well data from engineering domains can be represented in forms amenable to the underlying transformer architecture. For example, we expect that creative representations of time series data that leverage text or sequences will help unlock its potential for high-dimensional anomaly detection, fault diagnosis, and prognostics problems.

\subsection{Environmental Science and Sustainability}\label{sec:specifictasks:subsec:environmental_sustainability}

Given the multitude of urgent environmental challenges we face, the demand for innovative research methods to tackle these issues has become paramount. Some environmental fields using unstructured textual data, such as social surveys \cite{willcock2017comparison, marquart2019climate} or social media posts \cite{fox2020photosearcher, ghermandi2023social, fox2021enriching} can harness relatively simple \acp{LLM} methodologies such as sentiment analysis and topic clustering to enhance their research outputs. For example, the advent of \acp{LLM} has revolutionized how researchers can approach using social media data to address environmental challenges including, clean energy, climate change and ecosystem services \cite{kim2021public, uthirapathy2023topic, fan2023using}.

However, most environmental science research uses multi-modal data and diverse data sources, not just textual data. Here, \acp{LLM} can provide numerous unique opportunities for environmental science research using diverse data structures. For example, \acp{LLM} can be combined with image data and \ac{CV} methodologies for a range of tasks, including classifying satellite imagery \cite{roberts2023satin} or identifying species in photographs \cite{doi2023role}. 
\acp{LLM} can also help create simulations, including, generating underwater simulations for testing marine exploratory robots \cite{davinack2023can}, or ecosystem modeling for forecasting the impacts of environmental shifts on species distributions \cite{doi2023role}.

Environmental science research often requires interdisciplinary collaboration. \acp{LLM} can help to translate complex scientific concepts into language that is easier for experts and non-experts alike to understand, assisting in cross-disciplinary communication between researchers with different domain expertise \cite{agathokleous2023use}. Furthermore, researchers in environmental science can benefit from \acp{LLM} as they provide a common platform for sharing information and ideas, thus facilitating experts in different fields (as well as policy-makers and the general public), to collaborate and develop new strategies for pressing and holistic environmental challenges such as the impact of climate change on public health \cite{agathokleous2023use}.

\subsection{Health Sciences}\label{sec:specifictasks:subsec:health_sciences}

Health science studies deal with a range of textual data sets including patient information and notes, meta-analyses of previous studies, and social media discussions \cite{healthcare9091133}. However, simply using off-the-shelf \acp{LLM} for biomedical text mining frequently produces inadequate outcomes due to the lack of understanding of domain-specific terminology. Fine-tuned models have a more nuanced understanding of biomedical texts, thus providing state-of-the-art approaches to biomedical textual data analysis \cite{thirunavukarasu2023large}. Due to the rapidly increasing amount of biomedical documents, \acp{LLM} will be key to utilizing this data effectively. One example is BioBERT \cite{lee2020biobert}, which is the BERT model fine-tuned on biomedical texts and has shown significant improvement over off-the-shelf models in carrying out biomedical text-mining tasks.  Another \ac{LLM} specifically developed for health research is the GatorTron model \cite{yang2022large}, which was designed to improve clinical \ac{NLP} tasks. By increasing previous model parameters from 110 million to 8.9 billion parameters and improving the accuracy of clinical \ac{NLP} tasks, such as de-identification of records, the GatorTron model may pave the way for enabling medical AI systems to improve healthcare delivery.

\acp{LLM} can be an equally relevant tool for understanding problems of public health. For instance, the impact of the support of online communities in patients with mental disorders was assessed by \cite{healthcare9091133}, with the aid of language models for sentiment analysis of web scraped data. Analogously, \cite{fettachbenhiba} utilized topic modeling and social network analysis on Reddit data, in order to understand the relationship of Reddit communities and people suffering from eating disorders. With the help of NLP, \cite{antoniakmimnolevy} conduct an analysis of thousands of birth stories posted online, summarizing emotions and decisions taken in this often traumatic medical experience.

\subsection{Materials Science} 

\acp{LLM}, with their extraordinary text-mining capacities, can search through vast repositories of materials science literature, synthesizing vast amounts of information into actionable insights. This capability not only speeds up materials discovery, but also paves the way for novel connections between existing knowledge. However, several challenges remain: non-textual data, tacit knowledge, and jargon. To fully exploit the potential of \acp{LLM} in materials science, it is crucial to develop mechanisms that can bridge these gaps, such as developing domain-specific models, utilizing multimodal neural networks (which can combine text data and images \cite{radford2021learning}), and tightening collaboration within material science communities across the globe. 

Several groups developed materials-aware language models, namely MatSciBERT \cite{gupta2022matscibert}, MaterialBERT \cite{yoshitake2022materialbert}, and polymer-oriented MaterialsBERT \cite{shetty2023generalpurpose}. Unlike generic language models, these models were trained on a large corpus of materials science-centric publications, making them adept at understanding and generating content specifically in this domain. Such domain-specific \acp{LLM} can offer more accurate and contextually relevant outputs, substantially enhancing their utility for researchers in the field. For example, transformer architecture models were used for the prediction of metal–organic framework (MOF) synthesis \cite{zheng2023chatgpt} and property prediction \cite{cao2023moformer}. In comparison to structure-agnostic methods like Stoichiometric-120 and revised autocorrelation (RACs), a structure-agnostic deep learning method based on the transformer model, named as MOFormer, not only achieves significantly higher accuracy—21.4\% and 16.9\% better in band gap prediction and 35–48\% and 25–42\% better in various gas adsorption prediction tasks, respectively — but also outperforms the structure-based Smooth Overlap of Atomic Positions (SOAP) method in band gap prediction even with less training data \cite{cao2023moformer}. Additionally, pretraining has been shown to further enhance the model's performance, improving MOFormer's accuracy by an average of 5.34\% and 4.3\% for band gap and gas adsorption prediction, respectively \cite{cao2023moformer}.

Recently, collaborative effort across eight countries and 22 institutions set up a hackathon to delve into the \acp{LLM} for materials science \cite{jablonka2023examples}. This event underscored the growing interest and vast potential of \acp{LLM} within the materials science domain. Participants engaged in a comprehensive array of tasks, from accurate molecular energy predictions and prediction the compressive strength of concretes to molecule discovery by context and the extraction of insights from unstructured data sources. The variety of subjects tackled and the ability to produce working models in a short timeframe highlights the crucial role \acp{LLM} might play in the future of materials science. While the hackathon predominantly centered on materials science, the projects' diversity highlighted a salient feature of \acp{LLM}: their versatility and significant potential to build bridges connecting the diverse range of scientific disciplines. 

\subsection{Mathematics}
Despite the large number of \acp{LLM} publications in mathematics (as depicted in Fig. \ref{fig:mesh1}), upon closer inspection, the majority of the research articles listed fall under either applied fields that use mathematics or mathematical aspects of \acp{LLM}. As such, at the moment, \acp{LLM} are more of a subject of study in mathematics than a method applied to study problems in the field of mathematics. For example, the question of how to use \acp{LLM} for mathematical reasoning and problem-solving is in itself an active field of research \cite{showyourwork2021,stepbystep2023}, as some state-of-the-art \acp{LLM} when applied to commonsense logic and mathematics research have been shown to be not up to standard \cite{davis2023, mathematicalcapabilities2023}. Other directions of research go towards mathematical aspects of \acp{LLM} \cite{mathematicalhallucination2023,anthropic2021}. 

Perhaps due to the lack of theoretical understanding of \acp{LLM}, they have not been used in mathematics research. The interaction between mathematics and \acp{LLM}, for now, will be towards improving and understanding \acp{LLM}, and not using \acp{LLM} as serious generative tools to solve Mathematics problems, as seen in some other fields. However, this does not mean that \acp{LLM} have not entered the mathematics research pipeline through the tasks delineated in the previous section. Furthermore, for example, in computational mathematics and scientific computing there are already many examples of the use of \acp{LLM} in the computation of scientific problems (e.g. \cite{llmcovid2022,Kashefi2023}).

\subsection{Social Sciences} \label{sec:specifictasks:subsec:social_sciences}
The aforementioned work from \cite{korinek2023} provided a thorough guide on how \acp{LLM} can be used as a research assistant from the perspective of a social scientist. Beyond research assistantship, however, \acp{LLM} can be useful as a central tool in a myriad of Social Sciences' analysis, as indicated by Fig. \ref{fig:mesh1}, since data in text format is rather common in this scientific field. In this section, we discuss: (i) general methods involving \acp{LLM} useful for social scientists; (ii) \acp{LLM} as an object of social studies \textit{per se}; and (iii) several research examples employing language models in subfields ranging from Economics, Business, Finance, to technology adoption and public affairs.

\textbf{General \acp{LLM} tools}. Several methods employing \acp{LLM} can be potentially useful in the social scientist toolkit. The first one, already utilized by Environmental and Health Sciences (see sections \ref{sec:specifictasks:subsec:environmental_sustainability} and \ref{sec:specifictasks:subsec:health_sciences}), is sentiment analysis. \cite{Liu2015} provided learning materials for this task, which could employ \acp{LLM} or other language methods. Even though \acp{LLM} outperformed several tools for sentiment analysis, \cite{zhang2023sentiment} illuminated potential caveats of \acp{LLM} in tasks involving structured sentiment information. Another useful tool for the social scientist concerns the application of pre-trained models for specific language oriented tasks. For instance, \cite{barbieri-etal-2020-tweeteval} developed the TweetEval, a standardized tool to execute several tasks on tweets, including sentiment analysis, while also providing benchmarks for evaluating performance. A further distinct domain specific language model is SciBERT \cite{shen2022sscibert}, trained on the Social Science Citation Index journals. SciBERT performs state-of-the-art sequence tagging, sentence classification, dependency parsing, etc., in the academic literature of Social Sciences. A final tool worth mentioning regards the use of \acp{LLM} for aiding causal inference tasks. Notably, GPT-4 outperforms existing models in causal discovery tasks and counterfactual reasoning \cite{kıcıman2023causal}. In fields such as Social Sciences, Biostatistics, Computer Science, Epidemiology and Environmental Sciences, \acp{LLM} can aid researchers in creating Directed Acyclic Graphs (DAGs) that are essential for causal effects estimation. Researchers input e.g. book content, papers, or other sources of theoretical information into a language model, outputting the causal graph of interest \cite{long2023large}. Despite potential drawbacks, this application can drastically reduce the effort put into building causal graphs, or serving as a verification tool, in addition to existing methods in causal inference.

\textbf{\acp{LLM} being assessed}. \acp{LLM} themselves can be the object of study for social scientists. \cite{liu2023training} raised the concern that language models often fail to reflect societal norms and values in situations not seen before. These authors provided a methodology to train language models to better extrapolate learned values in new scenarios. Another set of researchers \cite{Binz_2023} applied a series of canonical cognitive psychology tests on GPT-3, finding that these models perform at least as good as humans in most tasks, but failing causal reasoning tasks. In related work, Deep Mind social researchers \cite{dasgupta2022language} tested language models on abstract reasoning problems. They found that the state-of-the-art \acp{LLM}, with 7-70 billion parameters, mimicked human behavior considerably closely in realistic situations, but performed poorly on abstract tasks disconnected from reality, which is in line with \cite{zhang2023understanding}.

\textbf{Sub-field specific applications}. 
\acp{LLM} were utilized to summarize a high volume of the Spanish parliament documents with RoBERTa and GPT-2 \cite{pena2023leveraging}, potentially making policy more straightforward to follow and aiding policy decision-making. \cite{jensenkarelltanigawalauhabashoudahfani2021} employed language models to analyze administrative data as well, measuring the spatial variation of religiosity in Indonesia. In the sub-fields of Business and Finance, \cite{DELLAROCAS200723} is a seminal work on sentiment analysis to forecast sales. More recently, \cite{su11030917} identified key factors on tweets that are correlated to startup business success. \cite{bollen2011} measured the correlation between moods captured on Twitter and the Dow Jones Industrial Average over time, while \cite{YADAV2020589} and \cite{lengkeek2023} performed sentiment analysis of financial news and micro-blogs, respectively. In the last month, \cite{zhang2023instructfingpt} demonstrated that although \acp{LLM} have been utilized for financial sentiment analysis, they still fail to accurately interpret numerical magnitudes and to capture some financial contexts. In terms of technology business, \cite{KAUFFMANN2020523} measured consumer sentiment towards high-tech products, and \cite{kwartengntsiful2020} and \cite{Caviggioli2020} established connections between technology adoption and their impact on consumers using language models. 

Finally, in the sub-field of Economics, \cite{hansenash2023} provides an overview of language-oriented tasks and methods for economists. \cite{NBERw31007} and \cite{chaubanabouvierfrank2023} studied the labor market with language models. While \cite{NBERw31007} analyzed job postings data with DistilBERT to measure the actual shift to remote work due to the COVID-19 pandemic, \cite{chaubanabouvierfrank2023} employed the Gensim language model to go through over one million higher education syllabi, establishing connections between higher education skills and workplace activities and earnings, using data from the Department of Labor. \cite{bajari2023hedonic} converted product text information to numeric data using \acp{LLM}, in order to measure hedonic prices, i.e. the contribution of products' characteristics to the final price of the product. A further application in the sub-field of Economics is rather speculative when compared to the applications described above, as it has yet to be fully explored. With data abundance, advancements in computing power, developments of refined ML and economic models, researchers are now able to estimate the effects of policy interventions or treatment effects for granular and specific subpopulations in the dataset \cite{atheywager2018, fogelmodenesi2023,jonathanheller2017}. Despite the granularity of results exhibited in these projects, characterizing these subpopulations at scale remains an outstanding challenge, especially when these relatively small subgroups lie at a complicated intersection of a multitude of variables -- e.g. gender, race, social economic status, education, etc. In order to tackle this subgroup characterization task, social scientists could employ \acp{LLM}, feeding into \acp{LLM} all of the characteristic values of each group -- e.g. a group could consist of mostly elderly Asian males -- outputting a subpopulation description from the language model. If successfully implemented, this could aid policy regulators and the general population to search for specific subgroups in a study, and understand their composition on the fly. In the limit, if researchers can feed the treatment effect estimates associated with each subgroup, \acp{LLM} could aid not only in explaining group dynamics, but aid in interpreting results and major trends that stem from complex models.

\section{Conclusion and Future Outlook}\label{sec:conclusion}
There has been much debate about the extent to which \acp{LLM} will catalyze a new research paradigm \cite{shen2023chatgpt,chacko2023paradigm}. Even in the short time that they have been readily available, pre-trained and fine-tuned \acp{LLM} are quickly becoming an indispensable tool for assisting in general research tasks. By automating research steps that are often manually carried out and time-intensive, such as information retrieval and code creation, language models will benefit most domains by increasing a researcher's efficiency and productivity. 

In Section 2, we discussed the immense potential of \acp{LLM} to facilitate ideation, coding, and writing tasks for researchers. Utilizing \acp{LLM} for these tasks also helps breakdown disciplinary barriers, especially when these barriers are maintained by jargon \cite{woodward2008more}. For example, \acp{LLM} can serve as powerful mediators in interdisciplinary dialogue, enabling researchers to communicate and share knowledge more efficiently. By unraveling domain-specific specialized terminology, they make work accomplished in one field available to all. This demystification of language encourages cross-disciplinary idea exchange, promoting a shared understanding of concepts. In this way, \acp{LLM} serve as bridges that connect researchers, fostering  interdisciplinary collaborations. As such, these models not only amplify the capabilities of researchers within their respective fields but also hold the promise of catalyzing interdisciplinary research \cite{agathokleous2023use}. In contrast, as we discussed in Section 3, when tasks are domain-specific or require high accuracy for data outside the training set, some form of fine-tuning will be necessary. For example, several fine-tuned BERT models have demonstrated the impact of fine-tuning to improve their efficiency for domain-specific information retrieval and review \cite{aeroBERT, SafeAeroBERT, astroBERT}. Furthermore, some domains are starting to investigate advanced \ac{NLP} tasks such as topic modeling and sentiment analysis of social media data for environmental science \cite{liu2019roberta}, and web scraping to incorporate online information to improve financial predictions \cite{lengkeek2023}. 

Here we have provided a current snapshot of \acp{LLM} and their applications, but recognizing they are still evolving is crucial. Existing literature has yet to push the boundaries of testing \acp{LLM} for solving fundamental science problems in research. These models cut across multiple disciplines, which utilize similar data types that are sequential but not necessarily textual. For example, \acp{LLM} hold immense potential to be trained to produce outputs that are not just deterministic responses but probabilistic inferences based on string representations of data. This capability could revolutionize the way we interpret and utilize data by enabling these models to estimate the likelihood of various outcomes or predict trends from raw data encoded as text. Such a development would allow \acp{LLM} to process and analyze large datasets, identify patterns, and generate statistical predictions, thereby serving as advanced tools for data analysis in fields ranging from scientific research to financial forecasting. The ability to output probabilities rather than certainties adds a layer of sophistication to the model's decision-making processes, reflecting more accurately the uncertain nature of real-world data and allowing for more nuanced and informed decision-making. Harnessing \acp{LLM} for the assessment of string representation data has already been explored in learning molecular properties \cite{ross_large-scale_2022}, the generation of new protein structures \cite{lin_evolutionary-scale_2023}, and the prediction of metal-organic framework properties \cite{cao2023moformer}. These innovative approaches show great promise for \acp{LLM} to revolutionize the research paradigm outside of \ac{NLP} or general research tasks. However, the applications of \acp{LLM} in this manner may not be ubiquitous across scientific domains and further testing is needed to validate its applicability for different domains. In contrast, Multimodal Language Models may provide the next major advancement in general use \acp{LLM}. These models can perceive input and generates output in arbitrary combinations (any-to-any) of text, image, video, and audio \cite{wu2023nextgpt}. For instance, these models may assist in the interpretation of complex figures, tables, and mathematics by processing them as input images and producing textual responses. By automating research steps that are often manually carried out and time-intensive, language models will benefit most domains by increasing a researcher's efficiency and productivity. 

Finally, it is important to consider that regardless of the application and despite their immense promise for streamlining the research workflow, there are several caveats that researchers need to acknowledge when harnessing \acp{LLM} for their research. As state-of-the-art \acp{LLM} are black box models, quantifying exactly how outputs are generated remains difficult. The results could be heavily influenced by biases \cite{politicalbias,biasneurips}, or hallucinations \cite{banghallucination,alkaissihallucination} arising from the training set. Ultimately, the researcher is responsible for their work, regardless of inaccuracies stemming from \ac{LLM} usage. Researchers need to be responsible when using \acp{LLM}, following the best possible ethical practices, and actively acknowledging and accounting for the limitations and biases of \acp{LLM} in their research.

This manuscript comprehensively outlines the dualistic nature of \acp{LLM} within an array of scientific disciplines, offering clarity on their distinct advantages and limitations. We have highlighted the universal strengths of \acp{LLM} in automating general tasks such as aggregating research data, refining manuscript development, and enhancing coding efficiency, which resonate across various scientific arenas. For example, in materials sciences, \acp{LLM} are capable in synthesizing vast quantities of data into actionable insights. In chemistry, they expedite the discovery and characterization of novel substances, markedly accelerating the research cycle.

Specific applications of \acp{LLM} are also discussed, with each field witnessing tailored benefits. In health sciences, \acp{LLM} demonstrate prowess in mining extensive datasets for patterns pertinent to diseases. \acp{LLM} in environmental science bolster climate modeling and biodiversity evaluations, essential for environmental stewardship. In engineering, these models have the potential to review safety regulations and foster security for workers, while in biological sciences, they advance the interpretation of taxonomic data, demanding precision to prevent the dissemination of inaccuracies.

Acknowledging these benefits, we do not overlook the need for ethical diligence, comprehensive verification measures, and expert involvement to ensure the responsible application of \acp{LLM}, thereby averting the propagation of biases and fallacies. We draw attention to the unequal accessibility of \acp{LLM} and call for transparency in their creation and operational aspects, the potential for misinformation and hallucinations, privacy concerns, lack of reproducibility of output as well as the consideration of their environmental implications.

Our evaluative commentary aims to foster further discussion and clear considerations of the integration of \acp{LLM} in scientific research, promoting a balanced view that capitalizes on their potential while conscientiously navigating their constraints. We conclude with a cautiously optimistic outlook, positing that \acp{LLM}, when employed judiciously, will indeed serve as a catalyst for more dynamic and introspective scientific inquiry.

\section{Acknowledgments}
This work is supported by the Eric and Wendy Schmidt AI in Science Postdoctoral Fellowship, a Schmidt Futures program.

\printbibliography

@article{wu2023nextgpt,
	title        = {{NExT-GPT: Any-to-Any Multimodal LLM}},
	author       = {Shengqiong Wu and Hao Fei and Leigang Qu and Wei Ji and Tat-Seng Chua},
	year         = 2023,
	journal      = {CoRR},
	volume       = {abs/2309.05519}
}

@misc{zhang2023understanding,
	title        = {{Understanding Causality with Large Language Models: Feasibility and Opportunities}},
	author       = {Cheng Zhang and Stefan Bauer and Paul Bennett and Jiangfeng Gao and Wenbo Gong and Agrin Hilmkil and Joel Jennings and Chao Ma and Tom Minka and Nick Pawlowski and James Vaughan},
	year         = 2023,
	eprint       = {2304.05524},
	archiveprefix = {arXiv},
	primaryclass = {cs.LG}
}

@misc{zhang2023instructfingpt,
	title        = {{Instruct-FinGPT: Financial Sentiment Analysis by Instruction Tuning of General-Purpose Large Language Models}},
	author       = {Boyu Zhang and Hongyang Yang and Xiao-Yang Liu},
	year         = 2023,
	eprint       = {2306.12659},
	archiveprefix = {arXiv},
	primaryclass = {cs.CL}
}

@article{YADAV2020589,
	title        = {{Sentiment analysis of financial news using unsupervised approach}},
	author       = {Anita Yadav and C K Jha and Aditi Sharan and Vikrant Vaish},
	year         = 2020,
	journal      = {Procedia Computer Science},
	volume       = 167,
	pages        = {589--598},
	doi          = {https://doi.org/10.1016/j.procs.2020.03.325},
	issn         = {1877-0509},
	url          = {https://www.sciencedirect.com/science/article/pii/S1877050920307912},
	note         = {International Conference on Computational Intelligence and Data Science}
}

@misc{showyourwork2021,
	title        = {{Show Your Work: Scratchpads for Intermediate Computation with Language Models}},
	author       = {Maxwell Nye and Anders Johan Andreassen and Guy Gur-Ari and Henryk Michalewski and Jacob Austin and David Bieber and David Dohan and Aitor Lewkowycz and Maarten Bosma and David Luan and Charles Sutton and Augustus Odena},
	year         = 2021,
	eprint       = {2112.00114},
	archiveprefix = {arXiv},
	primaryclass = {cs.LG}
}

@article{anthropic2021,
	title        = {{A Mathematical Framework for Transformer Circuits}},
	author       = {Elhage, Nelson and Nanda, Neel and Olsson, Catherine and Henighan, Tom and Joseph, Nicholas and Mann, Ben and Askell, Amanda and Bai, Yuntao and Chen, Anna and Conerly, Tom and DasSarma, Nova and Drain, Dawn and Ganguli, Deep and Hatfield-Dodds, Zac and Hernandez, Danny and Jones, Andy and Kernion, Jackson and Lovitt, Liane and Ndousse, Kamal and Amodei, Dario and Brown, Tom and Clark, Jack and Kaplan, Jared and McCandlish, Sam and Olah, Chris},
	year         = 2021,
	journal      = {Transformer Circuits Thread},
	note         = {https://transformer-circuits.pub/2021/framework/index.html}
}

@misc{davis2023,
	title        = {{Mathematics, word problems, common sense, and artificial intelligence}},
	author       = {Ernest Davis},
	year         = 2023,
	eprint       = {2301.09723},
	archiveprefix = {arXiv},
	primaryclass = {cs.AI}
}

@article{llmcovid2022,
	title        = {{GenSLMs: Genome-scale language models reveal SARS-CoV-2 evolutionary dynamics}},
	author       = {Maxim Zvyagin and Alexander Brace and Kyle Hippe and Yuntian Deng and Bin Zhang and Cindy Orozco Bohorquez and Austin Clyde and Bharat Kale and Danilo Perez-Rivera and Heng Ma and Carla M. Mann and Michael Irvin and J. Gregory Pauloski and Logan Ward and Valerie Hayot-Sasson and Murali Emani and Sam Foreman and Zhen Xie and Diangen Lin and Maulik Shukla and Weili Nie and Josh Romero and Christian Dallago and Arash Vahdat and Chaowei Xiao and Thomas Gibbs and Ian Foster and James J. Davis and Michael E. Papka and Thomas Brettin and Rick Stevens and Anima Anandkumar and Venkatram Vishwanath and Arvind Ramanathan},
	year         = 2022,
	journal      = {bioRxiv},
	publisher    = {Cold Spring Harbor Laboratory},
	doi          = {10.1101/2022.10.10.511571},
	url          = {https://www.biorxiv.org/content/early/2022/11/23/2022.10.10.511571},
	elocation-id = {2022.10.10.511571},
	eprint       = {https://www.biorxiv.org/content/early/2022/11/23/2022.10.10.511571.full.pdf}
}

@article{mathematicalhallucination2023,
	title        = {{A Mathematical Investigation of Hallucination and Creativity in GPT Models}},
	author       = {Lee, Minhyeok},
	year         = 2023,
	journal      = {Mathematics},
	volume       = 11,
	number       = 10,
	doi          = {10.3390/math11102320},
	issn         = {2227-7390},
	url          = {https://www.mdpi.com/2227-7390/11/10/2320},
	article-number = 2320
}

@misc{stepbystep2023,
	title        = {{Let's Verify Step by Step}},
	author       = {Hunter Lightman and Vineet Kosaraju and Yura Burda and Harri Edwards and Bowen Baker and Teddy Lee and Jan Leike and John Schulman and Ilya Sutskever and Karl Cobbe},
	year         = 2023,
	eprint       = {2305.20050},
	archiveprefix = {arXiv},
	primaryclass = {cs.LG}
}

@misc{mathematicalcapabilities2023,
	title        = {{Mathematical Capabilities of ChatGPT}},
	author       = {Simon Frieder and Luca Pinchetti and Alexis Chevalier and Ryan-Rhys Griffiths and Tommaso Salvatori and Thomas Lukasiewicz and Philipp Christian Petersen and Julius Berner},
	year         = 2023,
	eprint       = {2301.13867},
	archiveprefix = {arXiv},
	primaryclass = {cs.LG}
}

@techreport{hansenash2023,
	title        = {{Text Algorithms in Economics}},
	author       = {Hansen, Stephen and Ash, Elliott},
	year         = 2023,
	month        = Apr,
	number       = 18125,
	doi          = {},
	url          = {https://ideas.repec.org/p/cpr/ceprdp/18125.html},
	institution  = {C.E.P.R. Discussion Papers},
	type         = {CEPR Discussion Papers}
}

@misc{bajari2023hedonic,
	title        = {{Hedonic Prices and Quality Adjusted Price Indices Powered by AI}},
	author       = {Patrick Bajari and Zhihao Cen and Victor Chernozhukov and Manoj Manukonda and Suhas Vijaykumar and Jin Wang and Ramon Huerta and Junbo Li and Ling Leng and George Monokroussos and Shan Wan},
	year         = 2023,
	eprint       = {2305.00044},
	archiveprefix = {arXiv},
	primaryclass = {econ.GN}
}

@article{chaubanabouvierfrank2023,
	title        = {{Connecting higher education to workplace activities and earnings}},
	author       = {Chau, Hung and Bana, Sarah and Bouvier, Baptiste and Frank, Morgan},
	year         = 2023,
	journal      = {Plos one},
	volume       = 18,
	number       = 3,
	doi          = {https://doi.org/10.1371/journal.pone.0282323},
	url          = {https://journals.plos.org/plosone/article?id=10.1371/journal.pone.0282323}
}

@techreport{NBERw31007,
	title        = {{Remote Work across Jobs, Companies, and Space}},
	author       = {Hansen, Stephen and Lambert, Peter John and Bloom, Nicholas and Davis, Steven J and Sadun, Raffaella and Taska, Bledi},
	year         = 2023,
	month        = {03},
	series       = {Working Paper Series},
	number       = 31007,
	doi          = {10.3386/w31007},
	url          = {http://www.nber.org/papers/w31007},
	institution  = {National Bureau of Economic Research},
	type         = {Working Paper}
}

@inproceedings{barbieri-etal-2020-tweeteval,
	title        = {{TweetEval: Unified Benchmark and Comparative Evaluation for Tweet Classification}},
	author       = {Barbieri, Francesco  and Camacho-Collados, Jose  and Espinosa Anke, Luis  and Neves, Leonardo},
	year         = 2020,
	month        = nov,
	booktitle    = {Findings of the Association for Computational Linguistics: EMNLP 2020},
	publisher    = {Association for Computational Linguistics},
	address      = {Online},
	pages        = {1644--1650},
	doi          = {10.18653/v1/2020.findings-emnlp.148},
	url          = {https://aclanthology.org/2020.findings-emnlp.148}
}

@article{su11030917,
	title        = {{Detecting Indicators for Startup Business Success: Sentiment Analysis Using Text Data Mining}},
	author       = {Saura, Jose Ramon and Palos-Sanchez, Pedro and Grilo, Antonio},
	year         = 2019,
	journal      = {Sustainability},
	volume       = 11,
	number       = 3,
	doi          = {10.3390/su11030917},
	issn         = {2071-1050},
	url          = {https://www.mdpi.com/2071-1050/11/3/917},
	article-number = 917
}

@article{KAUFFMANN2020523,
	title        = {{A framework for big data analytics in commercial social networks: A case study on sentiment analysis and fake review detection for marketing decision-making}},
	author       = {Erick Kauffmann and Jesús Peral and David Gil and Antonio Ferrández and Ricardo Sellers and Higinio Mora},
	year         = 2020,
	journal      = {Industrial Marketing Management},
	volume       = 90,
	pages        = {523--537},
	doi          = {https://doi.org/10.1016/j.indmarman.2019.08.003},
	issn         = {0019-8501},
	url          = {https://www.sciencedirect.com/science/article/pii/S0019850118307612}
}

@article{Caviggioli2020,
	title        = {{Technology adoption news and corporate reputation: sentiment analysis about the introduction of Bitcoin}},
	author       = {Caviggioli, Federico and Lamberti, Lucio and Landoni, Paolo and Meola, Paolo},
	year         = 2020,
	month        = {01},
	day          = {01},
	journal      = {Journal of Product {\&} Brand Management},
	publisher    = {Emerald Publishing Limited},
	volume       = 29,
	number       = 7,
	pages        = {877--897},
	doi          = {10.1108/JPBM-03-2018-1774},
	issn         = {1061-0421},
	url          = {https://doi.org/10.1108/JPBM-03-2018-1774}
}

@inproceedings{kwartengntsiful2020,
	title        = {{Consumer Insight on Driverless Automobile Technology Adoption via Twitter Data: A Sentiment Analytic Approach}},
	author       = {Kwarteng, Michael Adu and Ntsiful, Alex and Botchway, Raphael Kwaku and Pilik, Michal and Oplatkov{\'a}, Zuzana Kom{\'i}nkov{\'a}},
	year         = 2020,
	booktitle    = {Re-imagining Diffusion and Adoption of Information Technology and Systems: A Continuing Conversation},
	publisher    = {Springer International Publishing},
	address      = {Cham},
	pages        = {463--473},
	isbn         = {978-3-030-64849-7},
	editor       = {Sharma, Sujeet K. and Dwivedi, Yogesh K. and Metri, Bhimaraya and Rana, Nripendra P.}
}

@article{bollen2011,
	title        = {{Twitter mood predicts the stock market}},
	author       = {Johan Bollen and Huina Mao and Xiaojun Zeng},
	year         = 2011,
	journal      = {Journal of Computational Science},
	volume       = 2,
	number       = 1,
	pages        = {1--8},
	doi          = {https://doi.org/10.1016/j.jocs.2010.12.007},
	issn         = {1877-7503},
	url          = {https://www.sciencedirect.com/science/article/pii/S187775031100007X}
}

@article{DELLAROCAS200723,
	title        = {{Exploring the value of online product reviews in forecasting sales: The case of motion pictures}},
	author       = {Chrysanthos Dellarocas and Xiaoquan (Michael) Zhang and Neveen F. Awad},
	year         = 2007,
	journal      = {Journal of Interactive Marketing},
	volume       = 21,
	number       = 4,
	pages        = {23--45},
	doi          = {https://doi.org/10.1002/dir.20087},
	issn         = {1094-9968},
	url          = {https://www.sciencedirect.com/science/article/pii/S1094996807700361}
}

@misc{Liu2015,
	title        = {{Sentiment analysis : mining opinions, sentiments, and emotions}},
	author       = {Liu, Bing},
	year         = 2015,
	publisher    = {Cambridge University Press New York, NY},
	address      = {New York, NY},
	language     = {eng}
}

@misc{dasgupta2022language,
	title        = {{Language models show human-like content effects on reasoning}},
	author       = {Ishita Dasgupta and Andrew K. Lampinen and Stephanie C. Y. Chan and Antonia Creswell and Dharshan Kumaran and James L. McClelland and Felix Hill},
	year         = 2022,
	eprint       = {2207.07051},
	archiveprefix = {arXiv},
	primaryclass = {cs.CL}
}

@article{Binz_2023,
	title        = {{Using cognitive psychology to understand GPT-3}},
	author       = {Marcel Binz and Eric Schulz},
	year         = 2023,
	month        = {02},
	journal      = {Proceedings of the National Academy of Sciences},
	publisher    = {Proceedings of the National Academy of Sciences},
	volume       = 120,
	number       = 6,
	doi          = {10.1073/pnas.2218523120},
	url          = {https://doi.org/10.1073%2Fpnas.2218523120}
}

@misc{shen2022sscibert,
	title        = {{SsciBERT: A Pre-trained Language Model for Social Science Texts}},
	author       = {Si Shen and Jiangfeng Liu and Litao Lin and Ying Huang and Lin Zhang and Chang Liu and Yutong Feng and Dongbo Wang},
	year         = 2022,
	eprint       = {2206.04510},
	archiveprefix = {arXiv},
	primaryclass = {cs.CL}
}

@misc{liu2023training,
	title        = {{Training Socially Aligned Language Models in Simulated Human Society}},
	author       = {Ruibo Liu and Ruixin Yang and Chenyan Jia and Ge Zhang and Denny Zhou and Andrew M. Dai and Diyi Yang and Soroush Vosoughi},
	year         = 2023,
	eprint       = {2305.16960},
	archiveprefix = {arXiv},
	primaryclass = {cs.CL}
}

@article{jensenkarelltanigawalauhabashoudahfani2021,
	title        = {{Language Models in Sociological Research: An Application to Classifying Large Administrative Data and Measuring Religiosity}},
	author       = {Jeffrey L. Jensen and Daniel Karell and Cole Tanigawa-Lau and Nizar Habash and Mai Oudah and Dhia Fairus Shofia Fani},
	year         = 2022,
	journal      = {Sociological Methodology},
	volume       = 52,
	number       = 1,
	pages        = {30--52},
	doi          = {10.1177/00811750211053370}
}

@misc{pena2023leveraging,
	title        = {{Leveraging Large Language Models for Topic Classification in the Domain of Public Affairs}},
	author       = {Alejandro Peña and Aythami Morales and Julian Fierrez and Ignacio Serna and Javier Ortega-Garcia and Iñigo Puente and Jorge Cordova and Gonzalo Cordova},
	year         = 2023,
	eprint       = {2306.02864},
	archiveprefix = {arXiv},
	primaryclass = {cs.AI}
}

@misc{AAS_editorial,
	title        = {{Editorial: On the Use of Chatbots in Writing Scientific Manuscripts}},
	note         = {Accessed: 2023-08-22},
	howpublished = {\url{https://baas.aas.org/pub/2023i016/release/1}}
}

@misc{ACS_editorial,
	title        = {{Best Practices for Using AI When Writing Scientific Manuscripts}},
	note         = {Accessed: 2023-08-22},
	howpublished = {\url{https://pubs.acs.org/doi/10.1021/acsnano.3c01544}}
}

@misc{elsevier_editorial,
	title        = {{Combustion and Flame: Guide for Authors}},
	note         = {Accessed: 2023-08-22},
	howpublished = {\url{https://www.elsevier.com/journals/combustion-and-flame/0010-2180/guide-for-authors}}
}

@misc{um_gpt,
	journal      = {U},
	url          = {https://genai.umich.edu/}
}

@misc{chat_gpt_enterprise,
	journal      = {Introducing ChatGPT Enterprise},
	url          = {https://openai.com/blog/introducing-chatgpt-enterprise}
}

@misc{nature_editorial,
	title        = {{Nature Editorial Policies: Artificial Intelligence}},
	note         = {Accessed: 2023-08-22},
	howpublished = {\url{https://www.nature.com/nature-portfolio/editorial-policies/ai}}
}

@article{southsecure,
	title        = {{Secure Community Transformers: Private Pooled Data for LLMs}},
	author       = {South, Tobin and Zuskind, Guy and Mahari, Robert and Hardjono, Thomas}
}

@article{bumgardner2023local,
	title        = {{Local Large Language Models for Complex Structured Medical Tasks}},
	author       = {Bumgardner, VK and Mullen, Aaron and Armstrong, Sam and Hickey, Caylin and Talbert, Jeff},
	year         = 2023,
	journal      = {arXiv preprint arXiv:2308.01727}
}

@misc{ICML_editorial,
	title        = {{ICML 2023: Clarification on Large Language Model Policy LLM}},
	note         = {Accessed: 2023-08-22},
	howpublished = {\url{https://icml.cc/Conferences/2023/llm-policy}}
}

@misc{science_editorial,
	title        = {{Science Journals: Editorial Policies}},
	note         = {Accessed: 2023-08-22},
	howpublished = {\url{https://www.science.org/content/page/science-journals-editorial-policies}}
}

@misc{OpenAI_privacy,
	journal      = {OpenAI},
	url          = {https://openai.com/policies/privacy-policy}
}

@article{2023arXiv230500108C,
	title        = {{A data science platform to enable time-domain astronomy}},
	author       = {{Coughlin}, Michael W. and {Bloom}, Joshua S. and {Nir}, Guy and {Antier}, Sarah and {Jegou du Laz}, Theophile and {van der Walt}, St{\'e}fan and {Crellin-Quick}, Arien and {Culino}, Thomas and {Duev}, Dmitry A. and {Goldstein}, Daniel A. and {Healy}, Brian F. and {Karambelkar}, Viraj and {Lilleboe}, Jada and {Shin}, Kyung Min and {Singer}, Leo P. and {Ahumada}, Tomas and {Anand}, Shreya and {Bellm}, Eric C. and {Dekany}, Richard and {Graham}, Matthew J. and {Kasliwal}, Mansi M. and {Kostadinova}, Ivona and {Weizmann Kiendrebeogo}, R. and {Kulkarni}, Shrinivas R. and {Jenkins}, Sydney and {LeBaron}, Natalie and {Mahabal}, Ashish A. and {Neill}, James D. and {Parazin}, B. and {Peloton}, Julien and {Perley}, Daniel A. and {Riddle}, Reed and {Rusholme}, Ben and {van Santen}, Jakob and {Sollerman}, Jesper and {Stein}, Robert and {Turpin}, Damien and {Wold}, Avery and {Amat}, Carla and {Bonnefon}, Adrien and {Bonnefoy}, Adrien and {Flament}, Manon and {Kerkow}, Frank and {Kishore}, Sulekha and {Jani}, Shloke and {Mahanty}, Stephen K. and {Liu}, C{\'e}line and {Llinares}, Laura and {Makarison}, Jolyane and {Olli{\'e}ric}, Alix and {Perez}, In{\`e}s and {Pont}, Lydie and {Sharma}, Vyom},
	year         = 2023,
	month        = apr,
	journal      = {arXiv e-prints},
	pages        = {arXiv:2305.00108},
	doi          = {10.48550/arXiv.2305.00108},
	eid          = {arXiv:2305.00108},
	archiveprefix = {arXiv},
	eprint       = {2305.00108},
	primaryclass = {astro-ph.IM},
	adsurl       = {https://ui.adsabs.harvard.edu/abs/2023arXiv230500108C}
}

@article{astroBERT,
	title        = {{Building astroBERT, a language model for Astronomy \& Astrophysics}},
	author       = {{Grezes}, Felix and {Blanco-Cuaresma}, Sergi and {Accomazzi}, Alberto and {Kurtz}, Michael J. and {Shapurian}, Golnaz and {Henneken}, Edwin and {Grant}, Carolyn S. and {Thompson}, Donna M. and {Chyla}, Roman and {McDonald}, Stephen and {Hostetler}, Timothy W. and {Templeton}, Matthew R. and {Lockhart}, Kelly E. and {Martinovic}, Nemanja and {Chen}, Shinyi and {Tanner}, Chris and {Protopapas}, Pavlos},
	year         = 2021,
	month        = dec,
	journal      = {arXiv e-prints},
	pages        = {arXiv:2112.00590},
	doi          = {10.48550/arXiv.2112.00590},
	eid          = {arXiv:2112.00590},
	archiveprefix = {arXiv},
	eprint       = {2112.00590},
	primaryclass = {cs.CL},
	adsurl       = {https://ui.adsabs.harvard.edu/abs/2021arXiv211200590G}
}

@misc{braşoveanu2022visualizing,
	title        = {{Visualizing and Explaining Language Models}},
	author       = {Adrian M. P. Braşoveanu and Răzvan Andonie},
	year         = 2022,
	eprint       = {2205.10238},
	archiveprefix = {arXiv},
	primaryclass = {cs.CL}
}

@article{agathokleous2023use,
	title        = {{Use of ChatGPT: What does it mean for biology and environmental science?}},
	author       = {Agathokleous, Evgenios and Saitanis, Costas J and Fang, Chao and Yu, Zhen},
	year         = 2023,
	journal      = {Science of The Total Environment},
	publisher    = {Elsevier},
	volume       = 888,
	pages        = 164154
}

@article{Kashefi2023,
	title        = {{ChatGPT for programming numerical methods}},
	author       = {Ali Kashefi and Tapan  Mukerji},
	year         = 2023,
	journal      = {Journal of Machine Learning for Modeling and Computing},
	volume       = 4,
	number       = 2,
	pages        = {1--74},
	issn         = {2689-3967}
}

@inproceedings{SafeAeroBERT,
	title        = {{SafeAeroBERT: Towards a Safety-Informed Aerospace-Specific Language Model}},
	shorttitle   = {{SafeAeroBERT}},
	author       = {Andrade, Sequoia R. and Walsh, Hannah S.},
	year         = 2023,
	month        = jun,
	booktitle    = {{AIAA} {AVIATION} 2023 {Forum}},
	publisher    = {American Institute of Aeronautics and Astronautics},
	address      = {San Diego, CA and Online},
	doi          = {10.2514/6.2023-3437},
	isbn         = {978-1-62410-704-7},
	url          = {https://arc.aiaa.org/doi/10.2514/6.2023-3437},
	urldate      = {2023-06-13},
	language     = {en}
}

@article{Lcking2021MultipleAF,
	title        = {{Multiple annotation for biodiversity: developing an annotation framework among biology, linguistics and text technology}},
	author       = {Andy L{\"u}cking and Christine Driller and Manuel Stoeckel and Giuseppe Abrami and Adrian Pachzelt and Alexander Mehler},
	year         = 2021,
	journal      = {Language Resources and Evaluation},
	volume       = 56,
	pages        = {807--855},
	url          = {https://api.semanticscholar.org/CorpusID:238784293}
}

@article{Thessen2012ApplicationsON,
	title        = {{Applications of Natural Language Processing in Biodiversity Science}},
	author       = {Anne E. Thessen and Hong Cui and Dmitry Y. Mozzherin},
	year         = 2012,
	journal      = {Advances in Bioinformatics},
	volume       = 2012,
	url          = {https://api.semanticscholar.org/CorpusID:2192224}
}

@inproceedings{Thessen2018AutomatedTE,
	title        = {{Automated Trait Extraction using ClearEarth, a Natural Language Processing System for Text Mining in Natural Sciences}},
	author       = {Anne E. Thessen and Jenette Preciado and Payoj Jain and James H. Martin and Martha Palmer and Riyaz Ahmad Bhat},
	year         = 2018,
	url          = {https://api.semanticscholar.org/CorpusID:64529700}
}

@misc{carbontracker_2020,
	title        = {{Carbontracker: Tracking and Predicting the Carbon Footprint of Training Deep Learning Models}},
	shorttitle   = {Carbontracker},
	author       = {Anthony, Lasse F. Wolff and Kanding, Benjamin and Selvan, Raghavendra},
	year         = 2020,
	month        = jul,
	publisher    = {arXiv},
	url          = {http://arxiv.org/abs/2007.03051},
	urldate      = {2023-06-23},
	note         = {arXiv:2007.03051 [cs, eess, stat]},
	language     = {en}
}

@article{antoniakmimnolevy,
	title        = {{Narrative Paths and Negotiation of Power in Birth Stories}},
	author       = {Antoniak, Maria and Mimno, David and Levy, Karen},
	year         = 2019,
	month        = 11,
	journal      = {Proc. ACM Hum.-Comput. Interact.},
	publisher    = {Association for Computing Machinery},
	address      = {New York, NY, USA},
	volume       = 3,
	number       = {CSCW},
	doi          = {10.1145/3359190},
	url          = {https://doi.org/10.1145/3359190},
	issue_date   = {2019-11},
	articleno    = 88,
	numpages     = 27
}

@misc{vaswani2017attention,
	title        = {{Attention Is All You Need}},
	author       = {Ashish Vaswani and Noam Shazeer and Niki Parmar and Jakob Uszkoreit and Llion Jones and Aidan N. Gomez and Lukasz Kaiser and Illia Polosukhin},
	year         = 2017,
	eprint       = {1706.03762},
	archiveprefix = {arXiv},
	primaryclass = {cs.CL}
}

@article{bills2023language,
	title        = {{Language models can explain neurons in language models}},
	author       = {Bills, Steven and Cammarata, Nick and Mossing, Dan and Tillman, Henk and Gao, Leo and Goh, Gabriel and Sutskever, Ilya and Leike, Jan and Wu, Jeff and Saunders, William},
	year         = 2023,
	journal      = {URL https://openaipublic. blob. core. windows. net/neuron-explainer/paper/index. html.(Date accessed: 14.05. 2023)}
}

@online{cnn-news-article,
	title        = {{FTC is investigating ChatGPT-maker OpenAI for potential harm to consumers}},
	author       = {Brian Fung},
	year         = 2023,
	month        = 7,
	url          = {https://www.cnn.com/2023/07/13/tech/ftc-openai-investigation/index.html},
	urldate      = {2023-08-21}
}

@article{pimentel,
	title        = {{Do Large Language Models Understand Chemistry? A Conversation with ChatGPT}},
	author       = {Castro Nascimento, Cayque Monteiro and Pimentel, André Silva},
	year         = 2023,
	journal      = {Journal of Chemical Information and Modeling},
	volume       = 63,
	number       = 6,
	pages        = {1649--1655},
	doi          = {10.1021/acs.jcim.3c00285},
	url          = {https://doi.org/10.1021/acs.jcim.3c00285},
	note         = {PMID: 36926868},
	eprint       = {https://doi.org/10.1021/acs.jcim.3c00285}
}

@misc{singh2023explaining,
	title        = {{Explaining black box text modules in natural language with language models}},
	author       = {Chandan Singh and Aliyah R. Hsu and Richard Antonello and Shailee Jain and Alexander G. Huth and Bin Yu and Jianfeng Gao},
	year         = 2023,
	eprint       = {2305.09863},
	archiveprefix = {arXiv},
	primaryclass = {cs.AI}
}

@misc{nytimesOpinionNoam,
	title        = {{Opinion | Noam Chomsky: The False Promise of ChatGPT --- nytimes.com}},
	author       = {Chomsky, Noam},
	year         = {},
	note         = {[Accessed 28-Jun-2023]},
	howpublished = {\url{https://www.nytimes.com/2023/03/08/opinion/noam-chomsky-chatgpt-ai.html}}
}

@misc{yue2023leveraging,
	title        = {{Leveraging LLMs for KPIs Retrieval from Hybrid Long-Document: A Comprehensive Framework and Dataset}},
	author       = {Chongjian Yue and Xinrun Xu and Xiaojun Ma and Lun Du and Hengyu Liu and Zhiming Ding and Yanbing Jiang and Shi Han and Dongmei Zhang},
	year         = 2023,
	eprint       = {2305.16344},
	archiveprefix = {arXiv},
	primaryclass = {cs.CL}
}

@misc{GoogleBard,
	journal      = {Google},
	publisher    = {Google},
	url          = {https://support.google.com/bard/answer/13594961?hl=en#your_data}
}

@article{jonathanheller2017,
	title        = {{Using Causal Forests to Predict Treatment Heterogeneity: An Application to Summer Jobs}},
	author       = {Davis, Jonathan M.V. and Heller, Sara B.},
	year         = 2017,
	month        = 5,
	journal      = {American Economic Review},
	volume       = 107,
	number       = 5,
	pages        = {546--50},
	doi          = {10.1257/aer.p20171000},
	url          = {https://www.aeaweb.org/articles?id=10.1257/aer.p20171000}
}

@article{doi2023role,
	title        = {{The role of large language models in ecology and biodiversity conservation: Opportunities and Challenges}},
	author       = {Doi, Hideyuki and Osawa, Takeshi and Tsutsumida, Narumasa},
	year         = 2023
}

@misc{FineTuningLLM,
	title        = {{Structured information extraction from complex scientific text with fine-tuned large language models}},
	author       = {Dunn, Alexander and Dagdelen, John and Walker, Nicholas and Lee, Sanghoon and Rosen, Andrew S. and Ceder, Gerbrand and Persson, Kristin and Jain, Anubhav},
	year         = 2022,
	month        = dec,
	publisher    = {arXiv},
	url          = {http://arxiv.org/abs/2212.05238},
	urldate      = {2023-06-13},
	note         = {arXiv:2212.05238 [cond-mat]},
	language     = {en}
}

@misc{kıcıman2023causal,
	title        = {{Causal Reasoning and Large Language Models: Opening a New Frontier for Causality}},
	author       = {Emre Kıcıman and Robert Ness and Amit Sharma and Chenhao Tan},
	year         = 2023,
	eprint       = {2305.00050},
	archiveprefix = {arXiv},
	primaryclass = {cs.AI}
}

@article{fan2023using,
	title        = {{Using Social Media Text Data to Analyze the Characteristics and Influencing Factors of Daily Urban Green Space Usage—A Case Study of Xiamen, China}},
	author       = {Fan, Chenjing and Li, Shiqi and Liu, Yuxin and Jin, Chenxi and Zhou, Lingling and Gu, Yueying and Gai, Zhenyu and Liu, Runhan and Qiu, Bing},
	year         = 2023,
	journal      = {Forests},
	publisher    = {MDPI},
	volume       = 14,
	number       = 8,
	pages        = 1569
}

@article{Lffler2020DatasetSI,
	title        = {{Dataset search in biodiversity research: Do metadata in data repositories reflect scholarly information needs?}},
	author       = {Felicitas L{\"o}ffler and Valentin Wesp and Birgitta K{\"o}nig-Ries and Friederike Klan},
	year         = 2020,
	journal      = {PLoS ONE},
	volume       = 16,
	url          = {https://api.semanticscholar.org/CorpusID:211532490}
}

@inproceedings{fettachbenhiba,
	title        = {{Pro-Eating Disorders and Pro-Recovery Communities on Reddit: Text and Network Comparative Analyses}},
	author       = {Fettach, Yousra and Benhiba, Lamia},
	year         = 2020,
	booktitle    = {Proceedings of the 21st International Conference on Information Integration and Web-Based Applications \& Services},
	location     = {Munich, Germany},
	publisher    = {Association for Computing Machinery},
	address      = {New York, NY, USA},
	series       = {iiWAS2019},
	pages        = {277–286},
	doi          = {10.1145/3366030.3366058},
	isbn         = 9781450371797,
	url          = {https://doi.org/10.1145/3366030.3366058},
	numpages     = 10
}

@article{flam-shepherd_language_2022,
	title        = {{Language models can learn complex molecular distributions}},
	author       = {Flam-Shepherd, Daniel and Zhu, Kevin and Aspuru-Guzik, Alán},
	year         = 2022,
	month        = jun,
	journal      = {Nat Commun},
	volume       = 13,
	number       = 1,
	pages        = 3293,
	doi          = {10.1038/s41467-022-30839-x},
	issn         = {2041-1723},
	url          = {https://www.nature.com/articles/s41467-022-30839-x},
	urldate      = {2023-08-22},
	copyright    = {2022 The Author(s)}
}

@article{fox2020photosearcher,
	title        = {{“photosearcher” package in R: An accessible and reproducible method for harvesting large datasets from Flickr}},
	author       = {Fox, Nathan and August, Tom and Mancini, Francesca and Parks, Katherine E and Eigenbrod, Felix and Bullock, James M and Sutter, Louis and Graham, Laura J},
	year         = 2020,
	journal      = {SoftwareX},
	publisher    = {Elsevier},
	volume       = 12,
	pages        = 100624
}

@article{fox2021enriching,
	title        = {{Enriching social media data allows a more robust representation of cultural ecosystem services}},
	author       = {Fox, Nathan and Graham, Laura J and Eigenbrod, Felix and Bullock, James M and Parks, Katherine E},
	year         = 2021,
	journal      = {Ecosystem Services},
	publisher    = {Elsevier},
	volume       = 50,
	pages        = 101328
}

@misc{Friedman_2023,
	title        = {{Our new promethean moment}},
	author       = {Friedman, Thomas L.},
	year         = 2023,
	month        = {03},
	journal      = {The New York Times},
	publisher    = {The New York Times},
	url          = {https://www.nytimes.com/2023/03/21/opinion/artificial-intelligence-chatgpt.html}
}

@article{ghermandi2023social,
	title        = {{Social media data for environmental sustainability: A critical review of opportunities, threats, and ethical use}},
	author       = {Ghermandi, Andrea and Langemeyer, Johannes and Van Berkel, Derek and Calcagni, Fulvia and Depietri, Yaella and Vigl, Lukas Egarter and Fox, Nathan and Havinga, Ilan and J{\"a}ger, Hieronymus and Kaiser, Nina and others},
	year         = 2023,
	journal      = {One Earth},
	publisher    = {Elsevier},
	volume       = 6,
	number       = 3,
	pages        = {236--250}
}

@article{gupta2022matscibert,
	title        = {{MatSciBERT: A materials domain language model for text mining and information extraction}},
	author       = {Gupta, T. and Zaki, M. and Krishnan, N.M.A. and others},
	year         = 2022,
	journal      = {npj Computational Materials},
	volume       = 8,
	pages        = 102,
	url          = {https://doi.org/10.1038/s41524-022-00784-w}
}

@article{Glen2022,
	title        = {{Natural language processing models that automate programming will transform chemistry research and teaching}},
	author       = {Hocky, Glen M. and White, Andrew D.},
	year         = 2022,
	journal      = {Digital Discovery},
	publisher    = {RSC},
	volume       = 1,
	pages        = {79--83},
	doi          = {10.1039/D1DD00009H},
	url          = {http://dx.doi.org/10.1039/D1DD00009H},
	issue        = 2
}

@article{jablonka2023examples,
	title        = {{14 Examples of How LLMs Can Transform Materials Science and Chemistry: A Reflection on a Large Language Model Hackathon}},
	author       = {Jablonka, K.M. and Ai, Q. and Al-Feghali, A. and Badhwar, S. and Bo-carsly, J.D. and Bran, A.M. and Bringuier, S. and Brinson, L.C. and Choudhary, K. and Circi, D. and Cox, S.},
	year         = 2023,
	journal      = {Digital Discovery}
}

@misc{devlin2019bert,
	title        = {{BERT: Pre-training of Deep Bidirectional Transformers for Language Understanding}},
	author       = {Jacob Devlin and Ming-Wei Chang and Kenton Lee and Kristina Toutanova},
	year         = 2019,
	eprint       = {1810.04805},
	archiveprefix = {arXiv},
	primaryclass = {cs.CL}
}

@techreport{fogelmodenesi2023,
	title        = {{What is a labor market? classifying workers and jobs using network theory.}},
	author       = {James Fogel and Bernardo Modenesi},
	year         = 2023
}

@article{Balhoff2013ASM,
	title        = {{A Semantic Model for Species Description Applied to the Ensign Wasps (Hymenoptera: Evaniidae) of New Caledonia}},
	author       = {James P. Balhoff and Istv{\'a}n Mik{\'o} and Matthew Jon Yoder and Patricia L. Mullins and Andrew R. Deans},
	year         = 2013,
	journal      = {Systematic Biology},
	volume       = 62,
	pages        = {639--659},
	url          = {https://api.semanticscholar.org/CorpusID:11483046}
}

@misc{liu-arxiv-2023,
	title        = {{Is Your Code Generated by ChatGPT Really Correct? Rigorous Evaluation of Large Language Models for Code Generation}},
	author       = {Jiawei Liu and Chunqiu Steven Xia and Yuyao Wang and Lingming Zhang},
	year         = 2023,
	eprint       = {2305.01210},
	archiveprefix = {arXiv},
	primaryclass = {cs.SE}
}

@article{kasneci2023chatgpt,
	title        = {{ChatGPT for good? On opportunities and challenges of large language models for education}},
	author       = {Kasneci, Enkelejda and Se{\ss}ler, Kathrin and K{\"u}chemann, Stefan and Bannert, Maria and Dementieva, Daryna and Fischer, Frank and Gasser, Urs and Groh, Georg and G{\"u}nnemann, Stephan and H{\"u}llermeier, Eyke and others},
	year         = 2023,
	journal      = {Learning and Individual Differences},
	publisher    = {Elsevier},
	volume       = 103,
	pages        = 102274
}

@article{kim2021public,
	title        = {{Public sentiment toward solar energy—opinion mining of twitter using a transformer-based language model}},
	author       = {Kim, Serena Y and Ganesan, Koushik and Dickens, Princess and Panda, Soumya},
	year         = 2021,
	journal      = {Sustainability},
	publisher    = {MDPI},
	volume       = 13,
	number       = 5,
	pages        = 2673
}

@inproceedings{biasneurips,
	title        = {{Bias Out-of-the-Box: An Empirical Analysis of Intersectional Occupational Biases in Popular Generative Language Models}},
	author       = {Kirk, Hannah Rose and Jun, Yennie and Volpin, Filippo and Iqbal, Haider and Benussi, Elias and Dreyer, Frederic and Shtedritski, Aleksandar and Asano, Yuki},
	year         = 2021,
	booktitle    = {Advances in Neural Information Processing Systems},
	publisher    = {Curran Associates, Inc.},
	volume       = 34,
	pages        = {2611--2624},
	url          = {https://proceedings.neurips.cc/paper_files/paper/2021/file/1531beb762df4029513ebf9295e0d34f-Paper.pdf},
	editor       = {M. Ranzato and A. Beygelzimer and Y. Dauphin and P.S. Liang and J. Wortman Vaughan}
}

@techreport{korinek2023,
	title        = {{Language Models and Cognitive Automation for Economic Research}},
	author       = {Korinek, Anton},
	year         = 2023,
	month        = {02},
	series       = {Working Paper Series},
	number       = 30957,
	doi          = {10.3386/w30957},
	url          = {http://www.nber.org/papers/w30957},
	institution  = {National Bureau of Economic Research},
	type         = {Working Paper}
}

@inproceedings{kulal-neurips-2019,
	title        = {{SPoC: Search-based Pseudocode to Code}},
	author       = {Kulal, Sumith and Pasupat, Panupong and Chandra, Kartik and Lee, Mina and Padon, Oded and Aiken, Alex and Liang, Percy S},
	year         = 2019,
	booktitle    = {Advances in Neural Information Processing Systems},
	publisher    = {Curran Associates, Inc.},
	volume       = 32,
	pages        = {},
	url          = {https://proceedings.neurips.cc/paper_files/paper/2019/file/7298332f04ac004a0ca44cc69ecf6f6b-Paper.pdf},
	editor       = {H. Wallach and H. Larochelle and A. Beygelzimer and F. d\textquotesingle Alch\'{e}-Buc and E. Fox and R. Garnett}
}

@article{lee2020biobert,
	title        = {{BioBERT: a pre-trained biomedical language representation model for biomedical text mining}},
	author       = {Lee, Jinhyuk and Yoon, Wonjin and Kim, Sungdong and Kim, Donghyeon and Kim, Sunkyu and So, Chan Ho and Kang, Jaewoo},
	year         = 2020,
	journal      = {Bioinformatics},
	publisher    = {Oxford University Press},
	volume       = 36,
	number       = 4,
	pages        = {1234--1240}
}

@article{healthcare9091133,
	title        = {{Why Do Users of Online Mental Health Communities Get Likes and Reposts: A Combination of Text Mining and Empirical Analysis}},
	author       = {Liu, Jingfang and Kong, Jun},
	year         = 2021,
	journal      = {Healthcare},
	volume       = 9,
	number       = 9,
	doi          = {10.3390/healthcare9091133},
	issn         = {2227-9032},
	url          = {https://www.mdpi.com/2227-9032/9/9/1133},
	article-number = 1133,
	pubmedid     = 34574907
}

@article{liu2019roberta,
	title        = {{Roberta: A robustly optimized bert pretraining approach}},
	author       = {Liu, Yinhan and Ott, Myle and Goyal, Naman and Du, Jingfei and Joshi, Mandar and Chen, Danqi and Levy, Omer and Lewis, Mike and Zettlemoyer, Luke and Stoyanov, Veselin},
	year         = 2019,
	journal      = {arXiv preprint arXiv:1907.11692}
}

@misc{luccioni_estimating_2022,
	title        = {{Estimating the Carbon Footprint of BLOOM, a 176B Parameter Language Model}},
	author       = {Luccioni, Alexandra Sasha and Viguier, Sylvain and Ligozat, Anne-Laure},
	year         = 2022,
	month        = nov,
	publisher    = {arXiv},
	url          = {http://arxiv.org/abs/2211.02001},
	urldate      = {2023-06-23},
	note         = {arXiv:2211.02001 [cs]},
	language     = {en}
}

@inproceedings{Nyamisa2017ASO,
	title        = {{A Survey of Information Retrieval Techniques}},
	author       = {Mang’are Fridah Nyamisa and Waweru Mwangi and Wilson K. Cheruiyot},
	year         = 2017
}

@article{Mora2018SemiautomaticEO,
	title        = {{Semi-automatic Extraction of Plants Morphological Characters from Taxonomic Descriptions Written in Spanish}},
	author       = {Mar{\'i}a A. Mora and Jos{\'e} Enrique Araya},
	year         = 2018,
	journal      = {Biodiversity Data Journal},
	url          = {https://api.semanticscholar.org/CorpusID:49662446}
}

@article{lengkeek2023,
	title        = {{Leveraging hierarchical language models for aspect-based sentiment analysis on financial data}},
	author       = {Matteo Lengkeek and Finn {van der Knaap} and Flavius Frasincar},
	year         = 2023,
	journal      = {Information Processing \& Management},
	volume       = 60,
	number       = 5,
	pages        = 103435,
	doi          = {https://doi.org/10.1016/j.ipm.2023.103435},
	issn         = {0306-4573},
	url          = {https://www.sciencedirect.com/science/article/pii/S0306457323001723}
}

@article{politicalbias,
	title        = {{More human than human: Measuring ChatGPT political bias}},
	author       = {Motoki, Fabio and Neto, Valdemar Pinho and Rodrigues, Victor},
	year         = 2023,
	journal      = {Public Choice},
	publisher    = {Springer},
	pages        = {1--21}
}

@article{Nguyen2019COPIOUSAG,
	title        = {{COPIOUS: A gold standard corpus of named entities towards extracting species occurrence from biodiversity literature}},
	author       = {Nhung T. H. Nguyen and Roselyn Gabud and Sophia Ananiadou},
	year         = 2019,
	journal      = {Biodiversity Data Journal},
	url          = {https://api.semanticscholar.org/CorpusID:59413507}
}

@inproceedings{nie2019combining,
	title        = {{Combining fact extraction and verification with neural semantic matching networks}},
	author       = {Nie, Yixin and Chen, Haonan and Bansal, Mohit},
	year         = 2019,
	booktitle    = {Proceedings of the AAAI Conference on Artificial Intelligence},
	volume       = 33,
	number       = {01},
	pages        = {6859--6866}
}

@misc{openai2023gpt4,
	title        = {{GPT-4 Technical Report}},
	author       = {OpenAI},
	year         = 2023,
	eprint       = {2303.08774},
	archiveprefix = {arXiv},
	primaryclass = {cs.CL}
}

@article{Vaithilingam-chi-2022,
	title        = {{Expectation vs. Experience: Evaluating the Usability of Code Generation Tools Powered by Large Language Models}},
	author       = {Priyan Vaithilingam and Tianyi Zhang and Elena L. Glassman},
	year         = 2022,
	journal      = {CHI Conference on Human Factors in Computing Systems Extended Abstracts}
}

@article{pursnani2023performance,
	title        = {{Performance of ChatGPT on the US Fundamentals of Engineering Exam: Comprehensive Assessment of Proficiency and Potential Implications for Professional Environmental Engineering Practice}},
	author       = {Pursnani, Vinay and Sermet, Yusuf and Demir, Ibrahim},
	year         = 2023,
	journal      = {arXiv preprint arXiv:2304.12198}
}

@online{reuters_gpt_users,
	title        = {{ChatGPT Sets Record for Fastest Growing User Base, Analyst Note}},
	author       = {Reuters},
	year         = 2023,
	month        = {02},
	day          = 1,
	url          = {https://www.reuters.com/technology/chatgpt-sets-record-fastest-growing-user-base-analyst-note-2023-02-01/},
	urldate      = {2023-06-15},
	organization = {Reuters}
}

@misc{thoppilan2022lamda,
	title        = {{LaMDA: Language Models for Dialog Applications}},
	author       = {Romal Thoppilan and Daniel De Freitas and Jamie Hall and Noam Shazeer and Apoorv Kulshreshtha and Heng-Tze Cheng and Alicia Jin and Taylor Bos and Leslie Baker and Yu Du and YaGuang Li and Hongrae Lee and Huaixiu Steven Zheng and Amin Ghafouri and Marcelo Menegali and Yanping Huang and Maxim Krikun and Dmitry Lepikhin and James Qin and Dehao Chen and Yuanzhong Xu and Zhifeng Chen and Adam Roberts and Maarten Bosma and Vincent Zhao and Yanqi Zhou and Chung-Ching Chang and Igor Krivokon and Will Rusch and Marc Pickett and Pranesh Srinivasan and Laichee Man and Kathleen Meier-Hellstern and Meredith Ringel Morris and Tulsee Doshi and Renelito Delos Santos and Toju Duke and Johnny Soraker and Ben Zevenbergen and Vinodkumar Prabhakaran and Mark Diaz and Ben Hutchinson and Kristen Olson and Alejandra Molina and Erin Hoffman-John and Josh Lee and Lora Aroyo and Ravi Rajakumar and Alena Butryna and Matthew Lamm and Viktoriya Kuzmina and Joe Fenton and Aaron Cohen and Rachel Bernstein and Ray Kurzweil and Blaise Aguera-Arcas and Claire Cui and Marian Croak and Ed Chi and Quoc Le},
	year         = 2022,
	eprint       = {2201.08239},
	archiveprefix = {arXiv},
	primaryclass = {cs.CL}
}

@inproceedings{roziere-neurips-2020,
	title        = {{Unsupervised Translation of Programming Languages}},
	author       = {Roziere, Baptiste and Lachaux, Marie-Anne and Chanussot, Lowik and Lample, Guillaume},
	year         = 2020,
	booktitle    = {Advances in Neural Information Processing Systems},
	publisher    = {Curran Associates, Inc.},
	volume       = 33,
	pages        = {20601--20611},
	url          = {https://proceedings.neurips.cc/paper_files/paper/2020/file/ed23fbf18c2cd35f8c7f8de44f85c08d-Paper.pdf},
	editor       = {H. Larochelle and M. Ranzato and R. Hadsell and M.F. Balcan and H. Lin}
}

@misc{shen2023chatgpt,
	title        = {{ChatGPT and other large language models are double-edged swords}},
	author       = {Shen, Yiqiu and Heacock, Laura and Elias, Jonathan and Hentel, Keith D and Reig, Beatriu and Shih, George and Moy, Linda},
	year         = 2023,
	journal      = {Radiology},
	publisher    = {Radiological Society of North America},
	volume       = 307,
	number       = 2,
	pages        = {e230163}
}

@misc{li2022explanations,
	title        = {{Explanations from Large Language Models Make Small Reasoners Better}},
	author       = {Shiyang Li and Jianshu Chen and Yelong Shen and Zhiyu Chen and Xinlu Zhang and Zekun Li and Hong Wang and Jing Qian and Baolin Peng and Yi Mao and Wenhu Chen and Xifeng Yan},
	year         = 2022,
	eprint       = {2210.06726},
	archiveprefix = {arXiv},
	primaryclass = {cs.CL}
}

@article{atheywager2018,
	title        = {{Estimation and Inference of Heterogeneous Treatment Effects using Random Forests}},
	author       = {Stefan Wager and Susan Athey},
	year         = 2018,
	journal      = {Journal of the American Statistical Association},
	publisher    = {Taylor & Francis},
	volume       = 113,
	number       = 523,
	pages        = {1228--1242},
	doi          = {10.1080/01621459.2017.1319839},
	url          = {https://doi.org/10.1080/01621459.2017.1319839},
	eprint       = {https://doi.org/10.1080/01621459.2017.1319839}
}

@misc{long2023large,
	title        = {{Can large language models build causal graphs?}},
	author       = {Stephanie Long and Tibor Schuster and Alexandre Piché and Department of Family Medicine and McGill University and Mila and Université de Montreal and ServiceNow Research},
	year         = 2023,
	eprint       = {2303.05279},
	archiveprefix = {arXiv},
	primaryclass = {cs.CL}
}

@article{thirunavukarasu2023large,
	title        = {{Large language models in medicine}},
	author       = {Thirunavukarasu, Arun James and Ting, Darren Shu Jeng and Elangovan, Kabilan and Gutierrez, Laura and Tan, Ting Fang and Ting, Daniel Shu Wei},
	year         = 2023,
	journal      = {Nature Medicine},
	publisher    = {Nature Publishing Group US New York},
	pages        = {1--11}
}

@article{aeroBERT,
	title        = {{aeroBERT-Classifier: Classification of Aerospace Requirements Using BERT}},
	shorttitle   = {{aeroBERT}-{Classifier}},
	author       = {Tikayat Ray, Archana and Cole, Bjorn F. and Pinon Fischer, Olivia J. and White, Ryan T. and Mavris, Dimitri N.},
	year         = 2023,
	month        = mar,
	journal      = {Aerospace},
	volume       = 10,
	number       = 3,
	pages        = 279,
	doi          = {10.3390/aerospace10030279},
	issn         = {2226-4310},
	url          = {https://www.mdpi.com/2226-4310/10/3/279},
	urldate      = {2023-06-13},
	language     = {en}
}

@article{uthirapathy2023topic,
	title        = {{Topic Modelling and Opinion Analysis On Climate Change Twitter Data Using LDA And BERT Model.}},
	author       = {Uthirapathy, Samson Ebenezar and Sandanam, Domnic},
	year         = 2023,
	journal      = {Procedia Computer Science},
	publisher    = {Elsevier},
	volume       = 218,
	pages        = {908--917}
}

@article{Vishniac2023Editorial,
	title        = {{Editorial: On the Use of Chatbots in Writing Scientific Manuscripts}},
	author       = {Vishniac, Ethan T.},
	year         = 2023,
	month        = {03},
	journal      = {Bulletin of the AAS},
	publisher    = {},
	note         = {https://baas.aas.org/pub/2023i016}
}

@article{Jeronymo2023InParsv2LL,
	title        = {{InPars-v2: Large Language Models as Efficient Dataset Generators for Information Retrieval}},
	author       = {Vitor Jeronymo and Luiz Henrique Bonifacio and Hugo Queiroz Abonizio and Marzieh Fadaee and Roberto de Alencar Lotufo and Jakub Zavrel and Rodrigo Nogueira},
	year         = 2023,
	journal      = {ArXiv},
	volume       = {abs/2301.01820}
}

@inproceedings{wang_smiles-bert_2019,
	title        = {{SMILES-BERT: Large Scale Unsupervised Pre-Training for Molecular Property Prediction}},
	shorttitle   = {{SMILES}-{BERT}},
	author       = {Wang, Sheng and Guo, Yuzhi and Wang, Yuhong and Sun, Hongmao and Huang, Junzhou},
	year         = 2019,
	month        = sep,
	booktitle    = {Proceedings of the 10th {ACM} {International} {Conference} on {Bioinformatics}, {Computational} {Biology} and {Health} {Informatics}},
	publisher    = {ACM},
	address      = {Niagara Falls NY USA},
	pages        = {429--436},
	doi          = {10.1145/3307339.3342186},
	isbn         = {978-1-4503-6666-3},
	url          = {https://dl.acm.org/doi/10.1145/3307339.3342186},
	urldate      = {2023-08-14}
}

@misc{zhang2023sentiment,
	title        = {{Sentiment Analysis in the Era of Large Language Models: A Reality Check}},
	author       = {Wenxuan Zhang and Yue Deng and Bing Liu and Sinno Jialin Pan and Lidong Bing},
	year         = 2023,
	eprint       = {2305.15005},
	archiveprefix = {arXiv},
	primaryclass = {cs.CL}
}

@article{white_future_2023,
	title        = {{The future of chemistry is language}},
	author       = {White, Andrew D.},
	year         = 2023,
	month        = jul,
	journal      = {Nat Rev Chem},
	volume       = 7,
	number       = 7,
	pages        = {457--458},
	doi          = {10.1038/s41570-023-00502-0},
	issn         = {2397-3358},
	url          = {https://www.nature.com/articles/s41570-023-00502-0},
	urldate      = {2023-07-11},
	copyright    = {2023 Springer Nature Limited}
}

@misc{zhao2019dynamic,
	title        = {{Dynamic Stale Synchronous Parallel Distributed Training for Deep Learning}},
	author       = {Xing Zhao and Aijun An and Junfeng Liu and Bao Xin Chen},
	year         = 2019,
	eprint       = {1908.11848},
	archiveprefix = {arXiv},
	primaryclass = {cs.DC}
}

@inproceedings{LLMCode-Xu,
	title        = {{A systematic evaluation of large language models of code}},
	author       = {Xu, Frank F. and Alon, Uri and Neubig, Graham and Hellendoorn, Vincent Josua},
	year         = 2022,
	month        = jun,
	booktitle    = {Proceedings of the 6th {ACM} {SIGPLAN} {International} {Symposium} on {Machine} {Programming}},
	publisher    = {ACM},
	address      = {San Diego CA USA},
	pages        = {1--10},
	doi          = {10.1145/3520312.3534862},
	isbn         = {978-1-4503-9273-0},
	url          = {https://dl.acm.org/doi/10.1145/3520312.3534862},
	urldate      = {2023-06-13},
	language     = {en}
}

@article{yang2022large,
	title        = {{A large language model for electronic health records}},
	author       = {Yang, Xi and Chen, Aokun and PourNejatian, Nima and Shin, Hoo Chang and Smith, Kaleb E and Parisien, Christopher and Compas, Colin and Martin, Cheryl and Costa, Anthony B and Flores, Mona G and others},
	year         = 2022,
	journal      = {NPJ Digital Medicine},
	publisher    = {Nature Publishing Group UK London},
	volume       = 5,
	number       = 1,
	pages        = 194
}

@article{zhu2023chatgpt,
	title        = {{ChatGPT and environmental research}},
	author       = {Zhu, Jun-Jie and Jiang, Jinyue and Yang, Meiqi and Ren, Zhiyong Jason},
	year         = 2023,
	journal      = {Environmental Science \& Technology},
	publisher    = {ACS Publications}
}

@article{ahmad_chemberta-2_2022,
	title        = {{ChemBERTa-2: Towards Chemical Foundation Models}},
	shorttitle   = {{ChemBERTa}-2},
	author       = {Ahmad, Walid and Simon, Elana and Chithrananda, Seyone and Grand, Gabriel and Ramsundar, Bharath},
	year         = 2022,
	doi          = {10.48550/ARXIV.2209.01712},
	url          = {https://arxiv.org/abs/2209.01712},
	urldate      = {2023-08-25},
	copyright    = {Creative Commons Attribution 4.0 International},
	note         = {Publisher: arXiv Version Number: 1}
}

@article{badini2023assessing,
	title        = {{Assessing the capabilities of ChatGPT to improve additive manufacturing troubleshooting}},
	author       = {Badini, Silvia and Regondi, Stefano and Frontoni, Emanuele and Pugliese, Raffaele},
	year         = 2023,
	journal      = {Advanced Industrial and Engineering Polymer Research},
	publisher    = {Elsevier}
}

@inproceedings{radford2021learning,
	title        = {{Learning transferable visual models from natural language supervision}},
	author       = {Radford, A. and Kim, J.W. and Hallacy, C. and Ramesh, A. and Goh, G. and Agarwal, S. and Sastry, G. and Askell, A. and Mishkin, P. and Clark, J. and Krueger, G.},
	year         = 2021,
	booktitle    = {International Conference on Machine Learning},
	pages        = {8748--8763},
	organization = {PMLR}
}

@article{ross_large-scale_2022,
	title        = {{Large-scale chemical language representations capture molecular structure and properties}},
	author       = {Ross, Jerret and Belgodere, Brian and Chenthamarakshan, Vijil and Padhi, Inkit and Mroueh, Youssef and Das, Payel},
	year         = 2022,
	month        = dec,
	journal      = {Nat Mach Intell},
	volume       = 4,
	number       = 12,
	pages        = {1256--1264},
	doi          = {10.1038/s42256-022-00580-7},
	issn         = {2522-5839},
	url          = {https://www.nature.com/articles/s42256-022-00580-7},
	urldate      = {2023-08-25}
}

@article{yoshitake2022materialbert,
	title        = {{Materialbert for natural language processing of materials science texts}},
	author       = {Yoshitake, M. and Sato, F. and Kawano, H. and Teraoka, H.},
	year         = 2022,
	journal      = {Sci. Technol. Adv. Mater.},
	volume       = 2,
	pages        = {372--380}
}

@article{shetty2023generalpurpose,
	title        = {{A general-purpose material property data extraction pipeline from large polymer corpora using natural language processing}},
	author       = {Shetty, P. and Rajan, AC. and Kuenneth, C. and Gupta, S. and Panchumarti, LP. and Holm, L. and Zhang, C. and Ramprasad, R.},
	year         = 2023,
	month        = {04},
	day          = 5,
	journal      = {npj Computational Materials},
	volume       = 9,
	number       = 1,
	pages        = 52
}

@article{zheng2023chatgpt,
	title        = {{ChatGPT Chemistry Assistant for Text Mining and Prediction of MOF Synthesis}},
	author       = {Zheng, Zhiling and Zhang, Oufan and Borgs, Christian and Chayes, Jennifer T and Yaghi, Omar M},
	year         = 2023,
	journal      = {arXiv preprint arXiv:2306.11296}
}

@article{cao2023moformer,
	title        = {{Moformer: self-supervised transformer model for metal--organic framework property prediction}},
	author       = {Cao, Zhonglin and Magar, Rishikesh and Wang, Yuyang and Barati Farimani, Amir},
	year         = 2023,
	journal      = {Journal of the American Chemical Society},
	publisher    = {ACS Publications},
	volume       = 145,
	number       = 5,
	pages        = {2958--2967}
}

@article{lin_evolutionary-scale_2023,
	title        = {{Evolutionary-scale prediction of atomic-level protein structure with a language model}},
	author       = {Lin, Zeming and Akin, Halil and Rao, Roshan and Hie, Brian and Zhu, Zhongkai and Lu, Wenting and Smetanin, Nikita and Verkuil, Robert and Kabeli, Ori and Shmueli, Yaniv and dos Santos Costa, Allan and Fazel-Zarandi, Maryam and Sercu, Tom and Candido, Salvatore and Rives, Alexander},
	year         = 2023,
	month        = mar,
	journal      = {Science},
	volume       = 379,
	number       = 6637,
	pages        = {1123--1130},
	doi          = {10.1126/science.ade2574},
	url          = {https://www.science.org/doi/10.1126/science.ade2574},
	urldate      = {2023-08-25}
}

@software{elicit,
	title        = {{Elicit: The AI Research Assistant}},
	author       = {{Ought}},
	year         = 2023,
	url          = {https://elicit.org},
	date         = {2023-02-22}
}

@software{bard2023,
	title        = {{Bard: A Large Language Model from Google AI}},
	author       = {Bard},
	year         = 2023,
	url          = {https://bard.google.com/}
}

@article{chacko2023paradigm,
	title        = {{Paradigm shift presented by Large Language Models (LLM) in Deep Learning}},
	author       = {Chacko, Neha and Chacko, Viju},
	year         = 2023,
	journal      = {ADVANCES IN EMERGING COMPUTING TECHNOLOGIES},
	publisher    = {Co-Text Publishers},
	pages        = 40
}

@misc{banghallucination,
	title        = {{A Multitask, Multilingual, Multimodal Evaluation of ChatGPT on Reasoning, Hallucination, and Interactivity}},
	author       = {Yejin Bang and Samuel Cahyawijaya and Nayeon Lee and Wenliang Dai and Dan Su and Bryan Wilie and Holy Lovenia and Ziwei Ji and Tiezheng Yu and Willy Chung and Quyet V. Do and Yan Xu and Pascale Fung},
	year         = 2023,
	eprint       = {2302.04023},
	archiveprefix = {arXiv},
	primaryclass = {cs.CL}
}

@article{alkaissihallucination,
	title        = {{Artificial hallucinations in ChatGPT: implications in scientific writing}},
	author       = {Alkaissi, Hussam and McFarlane, Samy I},
	year         = 2023,
	journal      = {Cureus},
	publisher    = {Cureus},
	volume       = 15,
	number       = 2
}

@article{woodward2008more,
	title        = {{More than just jargon--the nature and role of specialist language in learning disciplinary knowledge}},
	author       = {Woodward-Kron, Robyn},
	year         = 2008,
	journal      = {Journal of English for Academic Purposes},
	publisher    = {Elsevier},
	volume       = 7,
	number       = 4,
	pages        = {234--249}
}

@article{roberts2023satin,
	title        = {{Satin: A multi-task metadataset for classifying satellite imagery using vision-language models}},
	author       = {Roberts, Jonathan and Han, Kai and Albanie, Samuel},
	year         = 2023,
	journal      = {arXiv preprint arXiv:2304.11619}
}

@article{davinack2023can,
	title        = {{Can ChatGPT be leveraged for taxonomic investigations? Potential and limitations of a new technology}},
	author       = {DAVINACK, ANDREW A},
	year         = 2023,
	journal      = {Zootaxa},
	volume       = 5270,
	number       = 2,
	pages        = {347--350}
}

@article{willcock2017comparison,
	title        = {{A comparison of cultural ecosystem service survey methods within South England}},
	author       = {Willcock, Simon and Camp, Brittany J and Peh, Kelvin S-H},
	year         = 2017,
	journal      = {Ecosystem Services},
	publisher    = {Elsevier},
	volume       = 26,
	pages        = {445--450}
}

@article{marquart2019climate,
	title        = {{Climate change views, energy policy preferences, and intended actions across welfare state regimes: Evidence from the European Social Survey}},
	author       = {Marquart-Pyatt, Sandra T and Qian, Hui and Houser, Matthew K and McCright, Aaron M},
	year         = 2019,
	journal      = {International Journal of Sociology},
	publisher    = {Taylor \& Francis},
	volume       = 49,
	number       = 1,
	pages        = {1--26}
}
\end{document}